\documentclass[10pt,twocolumn,letterpaper]{article}

\usepackage{cvpr}
\usepackage{times}
\usepackage{epsfig}
\usepackage{graphicx}
\usepackage{amsmath}
\usepackage{amssymb}

\graphicspath{{./figs/}}

\usepackage{soul}
\usepackage{tabularx}

\newcommand{\norm}[1]{\left\| {#1} \right\|}

\renewcommand{\epsilon}{\varepsilon}
\def\utilde#1{\mathord{\vtop{\ialign{##\crcr
$\hfil\displaystyle{#1}\hfil$\crcr\noalign{\kern1.5pt\nointerlineskip}
$\hfil\tilde{}\hfil$\crcr\noalign{\kern1.5pt}}}}}
\usepackage[breaklinks=true,bookmarks=false]{hyperref}

\cvprfinalcopy %

\ifcvprfinal\fi
\begin{document}

\title{PolyTransform: Deep Polygon Transformer for Instance Segmentation}

\author{
	Justin Liang$^{1}$ \quad Namdar Homayounfar$^{1,2}$\\
	Wei-Chiu Ma$^{1,3}$ \quad Yuwen Xiong$^{1,2}$ \quad Rui Hu$^{1}$ \quad Raquel Urtasun$^{1,2}$\\
	$^{1}$Uber Advanced Technologies Group \quad $^{2}$University of Toronto \quad $^{3}$ MIT\\
	\small\texttt{\{justin.liang,namdar,weichiu,yuwen,rui.hu,urtasun\}@uber.com}
}

\maketitle
\thispagestyle{empty}

\begin{abstract}

In this paper, we propose PolyTransform, a novel instance segmentation algorithm that produces precise, geometry-preserving masks by combining the strengths of prevailing segmentation approaches and modern polygon-based methods. 
 In particular, we first exploit a segmentation network to generate  instance masks. We then convert the masks into a set of polygons that are then fed to a deforming network that transforms the polygons such that they better fit the object boundaries. 
 Our experiments on the challenging Cityscapes dataset show that our PolyTransform significantly improves the performance of the backbone instance segmentation network and ranks 1st on the Cityscapes test-set leaderboard. 
 We also show impressive gains in the  interactive annotation setting.
\end{abstract}

\vspace{-5mm}
\section{Introduction}
\vspace{-1mm}

The goal of {\it instance segmentation} methods is to identify all countable objects in the scene, and produce a mask for each of them. With the help of instance segmentation, we can have a better understanding of the scene \cite{zhou2017scene}, design robotics systems that are capable of complex manipulation tasks \cite{fazeli2019see}, and improve   perception systems of self-driving cars \cite{ma2019deep}. 
The task  is, however,  extremely challenging.
In comparison to the traditional semantic segmentation task that infers the category of each pixel in the image, instance segmentation also requires the system to have the extra notion of individual objects in order to associate each pixel with one of them. 
Dealing with the wide variability in the scale and appearance of objects as well as   occlusions and motion blur make this problem extremely difficult.   

\begin{figure*}[htb!]
\vspace{-5mm}
\includegraphics[width=0.98\linewidth,trim={0mm 0mm 0mm 0mm},clip]{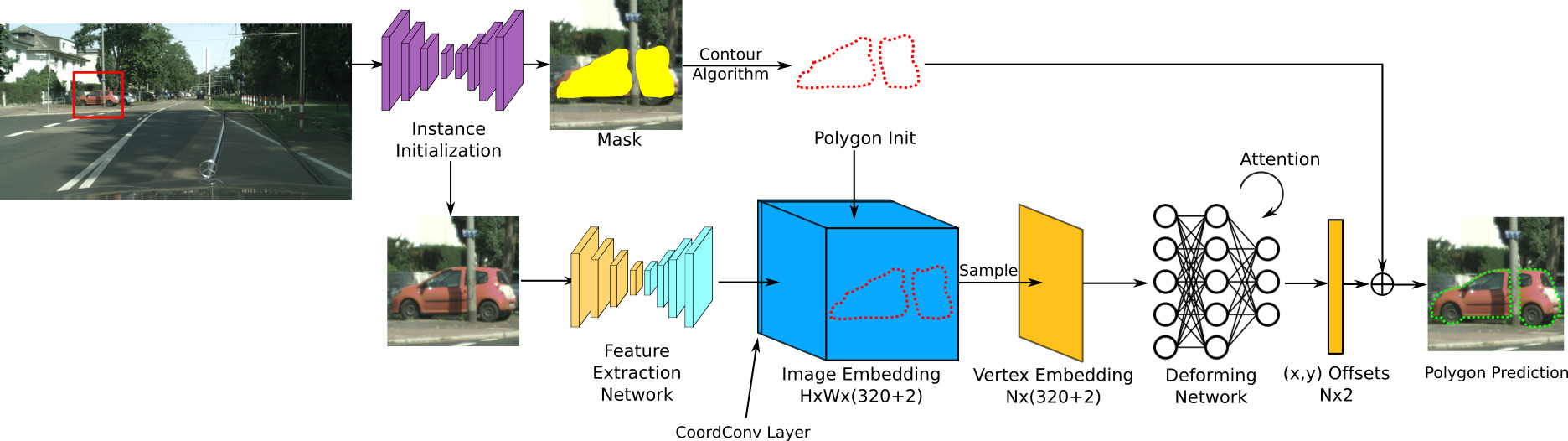}
\caption{\textbf{Overview of our PolyTransform model.}}
\vspace{-5mm}
\label{fig:network}
\end{figure*}

To address these issues, most modern instance segmentation methods employ a two-stage process \cite{mask-rcnn,upsnet,panet}, where  object proposals  are first created and  then   foreground background segmentation within each bounding box is performed. With the help of the box, they can better handle situations (\eg, occlusions) where other methods often fail \cite{bai2017deep}. 
While these approaches have achieved state-of-the-art performance on multiple benchmarks (e.g., COCO \cite{coco}, Cityscapes 
\cite{cityscapes}) their output is often over-smoothed failing to capture  fine-grained details. 
An alternative line of work tackles the problem of {\it interactive annotation} \cite{polygon-rnn,polygon-rnn++, wang2019delse, ling2019fast}. These techniques have been developed in the context of having an annotator in the loop, where a ground truth bounding box is provided. The goal of these works is to speed up annotation work by providing an initial polygon for annotators to correct as annotating from scratch is a very expensive process. In this line of work, methods exploit polygons to better capture the geometry of the object \cite{polygon-rnn,polygon-rnn++, ling2019fast}, instead of treating the problem as  a pixel-wise labeling task. This results in more precise masks and potentially faster annotation speed as annotators are able to simply correct the polygons by moving the vertices. 
However, these approaches suffer in the presence of  large occlusions or when the object is split into multiple disconnected components.  

With these problems in mind, in this paper we develop a novel model, which we call {\it PolyTransform}, and tackle both the {instance segmentation} and {interactive annotation} problems. The idea behind our approach is that the segmentation masks generated by common segmentation approaches can be viewed as a starting point to compute a set of  polygons, which can then  be refined. We performed this refinement via 
a deforming network that  predicts for each polygon the displacement of each vertex, taking into account the location of all vertices. By deforming each polygon, our model is able to better capture the local geometry of the object.   Unlike \cite{polygon-rnn,polygon-rnn++, ling2019fast},  our model has no restriction on the number of polygons utilized to describe each object. 
This allows us to naturally handle cases where the objects are split into parts due to occlusion. 

We first demonstrate the effectiveness of our approach on the Cityscapes dataset  \cite{cityscapes}.
On the task of instance segmentation, our model improves the initialization by \textbf{3.0} AP and \textbf{10.3} in the boundary metric on the validation set. Importantly, we achieve \textbf{1st} place on the test set leaderboard, beating the current state of the art by \textbf{3.7} AP. We further evaluate our model on a new self-driving dataset. Our model improves the initialization by \textbf{2.1} AP and \textbf{5.6} in the boundary metric.  In the context of interactive annotation, we outperform the previous state of the art \cite{wang2019delse} by \textbf{2.0}\% in the boundary metric. 
Finally, we conduct an experiment where the crowd-sourced labelers annotate the object instances using the polygon output from our model. We show that this can speed up the annotation time by \textbf{35}\%! 
\section{Related Work}
In this section, we briefly review the relevant literature on instance segmentation and annotation in the loop. 

\paragraph{Proposal-based Instance Segmentation:} 
Most modern instance segmentation models adopt a two-stage pipeline . First, an over-complete set   of segment proposals is identified, and then  a voting process is exploited to determine which one to keep \cite{Chen_2018_CVPR, Dai_2016_CVPR} %
As the explicit feature extraction process \cite{pont2017multiscale} is time-consuming \cite{hariharan2014simultaneous,hariharan2015hypercolumns}, Dai \etal \cite{dai2015convolutional,dai2016instance} integrated feature pooling into the neural network to improve  efficiency. While the speed is drastically boosted comparing to  previous approaches, it is still relatively slow as these approach is limited by the traditional detection based pipeline. With this problem in mind, researchers have looked into directly generating instance masks in the network and treat them as proposals \cite{pinheiro2015learning,pinheiro2016learning}. Based on this idea, Mask R-CNN \cite{mask-rcnn} introduced a joint approach to do both mask prediction and recognition. It builds upon Faster R-CNN \cite{ren2015faster} by adding an extra parallel header to predict the object's mask, in addition to the existing branch for bounding box recognition. Liu \etal \cite{panet} propose a path aggregation network to improve the information flow in Mask R-CNN  and further improve performance. More recently, Chen \etal  \cite{Chen_2019_CVPR} interleaves bounding box regression, mask regression and semantic segmentation together to boost instance segmentation performance. Xu \etal \cite{Xu_2019_ICCV} fit Chebyshev polynomials to instances by having a network learn the coefficients, this allows for real time instance segmentation. Huang \etal \cite{huang2019msrcnn} optimize the scoring of the bounding boxes by predicting IoU for each mask rather than only a classification score. Kuo \etal \cite{Kuo_2019_ICCV} start with bounding boxes and refine them using shape priors. Xiong \etal \cite{upsnet} and Kirillov \etal \cite{Kirillov_2019_CVPR} extended Mask R-CNN to the task of panoptic segmentation. Yang \etal \cite{Yang_2019_ICCV} extended Mask R-CNN to the task of video instance segmentation. %

\vspace{-0.2cm}
\paragraph{Proposal-free Instance Segmentation:} This line of research aims at segmenting the instances in the scene without an explicit object proposal. 
Zhang \etal \cite{zhang2015monocular,zhang2016instance} first predicts instance labels within the extracted multi-scale patches and then exploits dense Conditional Random Field \cite{krahenbuhl2011efficient} to obtain a consistent labeling of the full image. While achieving impressive results, their approach is computationally intensive. 
Bai and Urtasun \cite{bai2017deep} exploited a deep network to predict the energy of the watershed transform such that each basin corresponds to an object instance. With one simple cut, they can obtain the instance masks of the whole image without any post-processing. Similarly, \cite{kirillov2017instancecut} exploits boundary prediction to separate the instances within the same semantic category. Despite being much faster, they suffer when dealing with far or small objects whose boundaries are ambiguous.
To address this issue, Liu \etal \cite{liu2017sgn} present a sequential grouping approach that employs neural networks to gradually compose objects from simpler elements. It can robustly handle situations where a single instance is split into multiple parts. Newell and Deng \cite{newell2017associative} implicitly encode the grouping concept into neural networks by having the model to predict both semantic class and a tag for each pixel. The tags are one dimensional embeddings which associate each pixel with one another. Kendall \etal \cite{Kendall_2018_CVPR} propose a method to assign pixels to objects having each pixel point to its object's center so that it can be grouped. Sofiiuk \etal  \cite{Sofiiuk2019AdaptISAI} use a point proposal network to generate points where the instances can be, this is then processed by a CNN to outputs instance masks for each location. Neven \etal \cite{Neven_2019_CVPR} propose a new clustering loss that pulls the spatial embedding of pixels belonging to the same instance together to achieve real time instance segmentation while having high accuracy. Gao \etal \cite{Gao_2019_ICCV} propose a single shot instance segmentation network that outputs a pixel pair affinity pyramid to compute whether two pixels belong to the same instance, they then combine this with a predicted semantic segmentation to output a single instance segmentation map. %

\vspace{-0.2cm}

\paragraph{Interactive Annotation:} The task of interactive annotation can also be posed as finding the polygons or curves that best fit the object boundaries. In fact, the concept of deforming a curve to fit the object contour can be dated back to the 80s where the active contour model was first introduced \cite{kass1988snakes}. Since then,  variants of ACM \cite{cohen1991active,marcos2018learning,dom2019darnet} have been proposed to better capture the shape.  %
Recently, the idea of exploiting polygons to represent an instance is explored in the context of human in the loop segmentation \cite{polygon-rnn,polygon-rnn++}. Castrej{\'o}n \etal \cite{polygon-rnn} adopted an RNN to sequentially predict the vertices of the polygon. Acuna \etal \cite{polygon-rnn++} extended \cite{polygon-rnn} by incorporating graph neural networks and increasing image resolution. While these methods demonstrated promising results on public benchmarks \cite{cityscapes}, they require ground truth bounding box as input. Ling \etal \cite{ling2019fast} and Dong \etal \cite{deep-active-curve} exploited splines as an alternative parameterization. Instead of drawing the whole polygon/curves from scratch, they start with a circle and deform it. Wang \etal  tackled this problem with implicit curves using level sets \cite{wang2019delse}, however, because the outputs are not polygons, an annotator cannot easily corrected them. In \cite{dextr}, Maninis \etal use extreme boundary as inputs rather than bounding boxes and Majumder \etal \cite{Majumder_2019_CVPR} uses user clicks to generate content aware guidance maps; all of these help the networks learn stronger cues to generate more accurate segmentations. However, because they are pixel-wise masks, they are not easily amenable by an annotator. Acuna \etal \cite{AcunaCVPR19STEAL} develop an approach that can be used to refine noisy annotations by jointly reasoning about the object boundaries with a CNN and a level set formulation. In the domain of offline mapping, several papers from Homayounfar \etal and Liang \etal \cite{HomayounfarMLU18, LiangHMWU19, Homayounfar_2019_ICCV, Liang2018EndtoEndDS} have tackled the problem of automatically annotating crosswalks, road boundaries and lanes by predicting structured outputs such as a polyline.

\begin{figure}[tb!]
\includegraphics[width=0.98\linewidth,trim={0mm 0mm 0mm 0mm},clip]{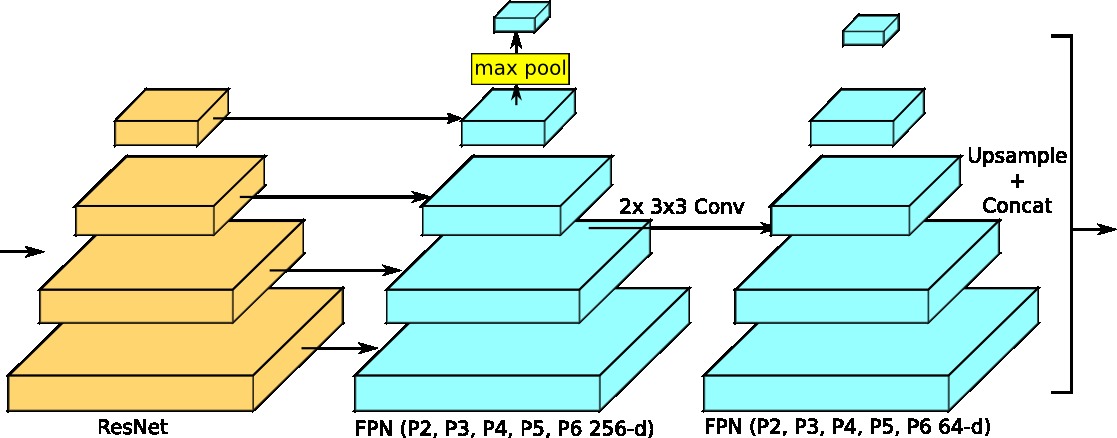}
\caption{\textbf{Our feature extraction network.}}
\vspace{-4mm}
\label{fig:fpn}
\end{figure}

\section{PolyTransform}
Our aim is to design a robust segmentation model that is capable of producing precise, geometry-preserving masks for each individual object. 
Towards this goal, we develop PolyTransform, a novel deep architecture that combines prevailing segmentation approaches \cite{mask-rcnn,upsnet} with modern polygon-based methods \cite{polygon-rnn,polygon-rnn++}. By exploiting the best of both worlds, we are able to generate high quality segmentation masks under various challenging scenarios.

In this section, we start by describing the backbone architecture for feature extraction and polygon initialization. 
Next, we present a novel deforming network that warps the initial polygon to better capture the local geometry of the object. An overview of our approach is shown in Figure \ref{fig:network}.

\subsection{Instance Initialization}
The goal of our instance initialization module is to provide a good polygon initialization for each individual object. To this end, we first exploit a  model to generate a mask for each instance in the scene. Our experiments show that our approach can significantly improve performance for a wide variety of segmentation models.  If the segmentation model outputs proposal boxes, we use them to crop the image, otherwise, we fit a bounding box to the mask. The cropped image is later resized to a square and fed into a feature network (described in Sec. \ref{sec:fpn})
to obtain a set of reliable deep features. In practice, we resize the cropped image to $(H_c, W_c) = (512, 512)$. To initialize the polygon, we use the border following algorithm of \cite{border-follow} to extract the contours from the predicted mask. We get the initial set of vertices by placing a vertex at every $10$ px distance in the contour. 
Empirically, we find such dense vertex interpolation provides a good balance between performance and memory consumption.

\subsection{Feature Extraction Network}
\label{sec:fpn}
The goal of our feature extraction network is to learn strong object boundary features. This is essential as we want our polygons to capture high curvature and complex shapes. %
As such, we employ a feature pyramid network (FPN) \cite{fpn} to learn and make use of multi-scale features. This network takes as input the $(H_c, W_c)$ crop obtained from the instance initialization stage and outputs a set of features at different pyramid levels. Our backbone can be seen in Figure \ref{fig:fpn}.

\setlength{\tabcolsep}{2pt}

\begin{table*}[t!]
\vspace{-0.5cm}
\setlength{\tabcolsep}{2pt}
\centering
  \begin{tabular}{|c|c|c|cc|cccccccc|}
  \cline{2-13}
  \multicolumn{1}{c|}{} & \multicolumn{1}{c|}{training data} &  \multicolumn{1}{c|}{AP$_ {\texttt{val}}$} & \multicolumn{1}{c}{AP} & AP$_{50}$ & \multicolumn{1}{c}{person}& \multicolumn{1}{c}{rider} & \multicolumn{1}{c}{car} & \multicolumn{1}{c}{truck} & \multicolumn{1}{c}{bus} & \multicolumn{1}{c}{train} & \multicolumn{1}{c}{mcycle} & bcycle \\ 
  \hline

  DWT \cite{bai2017deep}  & \texttt{fine} &$21.2$ &$19.4$ &$35.3$ &$15.5$ &$14.1$ &$31.5$ &$22.5$ &$27.0$ &$22.9$ &$13.9$ &$8.0$ \\
  Kendall et al. \cite{Kendall_2018_CVPR}  & \texttt{fine} &$-$ &$21.6$ &$39.0	$ &$19.2$ &$21.4$ &$36.6$ &$18.8$ &$26.8$ &$15.9$ &$19.4$ &$14.5$ \\
  Arnab et al. \cite{ArnabT17}  & \texttt{fine}  &$-$ &$23.4$ &$45.2$ &$21.0$ &$18.4$ &$31.7$ &$22.8$ &$31.1$ &$31.0$ &$19.6$ &$11.7$\\
  SGN \cite{liu2017sgn}  & \texttt{fine+coarse} &$29.2$ &$25.0$ &$44.9$ 
    &$21.8$ &$20.1$ &$39.4$ &$24.8$ &$33.2$ &$30.8$ &$17.7$ &$12.4$ \\
  PolygonRNN++ \cite{polygon-rnn++}  & \texttt{fine} &$-$ &$25.5$ &$45.5$ &$29.4$ &$21.8$ &$48.3$ &$21.2$ &$32.3$ &$23.7$ &$13.6$ &$13.6$ \\
  Mask R-CNN \cite{mask-rcnn} & \texttt{fine} &$31.5$ &$26.2$ &$49.9$ &$30.5$ &$23.7$ &$46.9$ &$22.8$ &$32.2$ &$18.6$ &$19.1$ &$16.0$ \\
  BShapeNet+ \cite{bshapenet} & \texttt{fine} & $-$ &$27.3$ &$50.4$
    &$29.7$ &$23.4$ &$46.7$ &$26.1$ &$33.3$ &$24.8$ &$20.3$ &$14.1$ \\
  GMIS  \cite{Liu2018AffinityDA}  & \texttt{fine+coarse} &$-$  & $27.3$ &$ 45.6$ &$31.5 $ &$25.2$ &$ 42.3 $ &$21.8$ &$ 37.2$ &$ 28.9$ &$18.8$ &$ 12.8$ \\
  Neven et al.  \cite{Neven_2019_CVPR}  & \texttt{fine} &$-$  & $27.6$ &$ 50.9$ &$34.5$ &$26.1$ &$ 52.4$ &$ 21.7$ &$ 31.2$ &$ 16.4$ &$ 20.1$ &$ 18.9$ \\
  PANet  \cite{panet}  & \texttt{fine} & $36.5$ &$31.8$ &$57.1$ &$36.8$ &$30.4$ &$54.8$ &$27.0$ &$36.3$ &$25.5$ &$22.6$ &$20.8$ \\

  Mask R-CNN \cite{mask-rcnn}  & \texttt{fine+COCO} &$36.4$ &$32.0$ &$58.1$ &$34.8$ &$27.0$ &$49.1$ &$30.1$ &$40.9$ &$30.9$ &$24.1$ &$18.7$ \\
   AdaptIS  \cite{Sofiiuk2019AdaptISAI}  & \texttt{fine} & $36.3$ &$32.5$ &$52.5$ & 31.4 &29.1 &50.0 &31.6 &41.7&39.4 &$24.7$ &12.1 \\
  SSAP \cite{Gao_2019_ICCV}  & \texttt{fine} &$37.3$ &$32.7$ &$51.8$ 
    &$35.4$ &$25.5$ &$55.9$ &$ 33.2$ &$43.9$ &$31.9$ &$19.5$ &$16.2$ \\
  BShapeNet+ \cite{bshapenet} & \texttt{fine+COCO} & $-$ &$32.9$ &$58.8$ 
    &$36.6$ &$24.8$ &$50.4$ &$33.7$ &$41.0$ &$33.7$ &$25.4$ &$17.8$ \\
  UPSNet \cite{upsnet} & \texttt{fine+COCO} &$37.8$ &$33.0$ &$59.7$ &$35.9$ &$27.4$ &$51.9$ &$31.8$ &$43.1$ &$31.4$ &$23.8$ &$19.1$ \\
  PANet \cite{panet} & \texttt{fine+COCO} &$41.4$ &$36.4$ &$63.1$ 
    &$41.5$ &$33.6$ &$58.2$ &$31.8$ &$45.3$ &$28.7$ &$28.2$ &$\bf{24.1}$ \\

  Ours & \texttt{fine+COCO} &$\bf{44.6}$ & $\bf{40.1}$ & $\bf{65.9}$ & $\bf{42.4}$ & $\bf{34.8}$ & $\bf{58.5}$ & $\bf{39.8}$ & $\bf{50.0}$ & $\bf{41.3}$ & $\bf{30.9}$ & $23.4$\\
  \hline

  \end{tabular}\\
  \caption{\textbf{Instance segmentation on Cityscapes val and test set:} This table shows our instance segmentation results on Cityscape test. We report models trained on \texttt{fine} and \texttt{fine+COCO}. We report AP and AP$_{50}$.}
  \label{tab:inst-test-results}
  \vspace{-1mm}
\end{table*}

\setlength{\tabcolsep}{2.25pt}
\begin{table*}[t!]
\centering
  \begin{tabular}{|c|c|c|cc|cccccccccc|}
  \cline{2-15}
  \multicolumn{1}{c|}{} & \texttt{fine} & \texttt{COCO} & \multicolumn{1}{c}{AP} & AP$_{50}$ & \multicolumn{1}{c}{car}& \multicolumn{1}{c}{truck} & \multicolumn{1}{c}{bus} & \multicolumn{1}{c}{train} & \multicolumn{1}{c}{person} & \multicolumn{1}{c}{rider} & \multicolumn{1}{c}{bcycle+r} & \multicolumn{1}{c}{bcycle} & \multicolumn{1}{c}{mcycle+r} & \multicolumn{1}{c|}{mcycle} \\ 
  \hline 
    
  Mask RCNN \cite{mask-rcnn} & \checkmark & - &$26.6$ &$53.5$  
    &$47.0$ &$41.1$ &$42.8$ &$10.7$ &$32.8$ &$27.5$ &$18.6$ &$10.2$  &$14.8$ &$20.2$ \\
  PANet \cite{panet} & \checkmark & - &$26.6$ &$53.5$  
    &$46.6$ &$41.8$ &$44.2$ &$2.7$ &$32.8$ &$27.4$ &$18.7$ &$11.3$  &$15.1$ &$25.8$ \\
  UPSNet \cite{upsnet} & \checkmark & - &$29.0$ &$56.0$  
    &$47.1$ &$41.8$ &$47.8$ &$12.7$ &$33.5$ &$27.3$ &$18.6$ &$10.4$  &$20.4$ &$30.2$ \\
  PANet \cite{panet} & \checkmark & \checkmark &$29.1$ &$55.2$  
    &$47.4$ &$43.7$ &$47.6$ &$10.7$ &$34.4$ &$30.1$ &$20.5$ &$11.8$  &$17.3$ &$27.4$ \\  
  UPSNet \cite{upsnet} & \checkmark & \checkmark &$31.5$ &$58.4$  
    &$46.9$ &$44.0$ &$49.8$ &$21.6$ &$34.1$ &$30.3$ &$21.7$ &$12.8$  &$19.3$  &$\bf{34.5}$ \\
  Ours & \checkmark & \checkmark &$\bf{35.3}$ &$\bf{60.8}$  
    &$\bf{50.5}$ &$\bf{47.3}$ &$\bf{52.5}$ &$\bf{23.4}$ &$\bf{40.4}$ &$\bf{37.0}$ &$\bf{25.1}$ &$\bf{16.0}$  &$\bf{28.7}$  &$32.6$ \\
  \hline

  \end{tabular}\\
  \caption{\textbf{Instance segmentation on test set of our new self-driving dataset:} This table shows our instance segmentation results our new dataset's test set. We report models trained on \texttt{fine} and \texttt{fine+COCO}. We report AP and AP$_{50}$. +r is short for with rider.}
  \label{tab:inst-test-results-uber}
  \vspace{-3mm}
\end{table*}

\subsection{Deforming Network}
We have   computed a polygon initialization and deep features of the FPN from the image crop. %
Next  we build a feature embedding for all $N$ vertices and learn a deforming model that can effectively predict the offset for each vertex so that the polygon snaps better to the object boundaries.%

\paragraph{Vertex embedding:}
We build our vertex representation upon the multi-scale feature extracted from the backbone FPN network of the previous section. In particular, we take the $P_2$, $P_3$, $P_4$, $P_5$ and $P_6$ feature maps and apply two lateral convolutional layers to each of them in order to reduce the number of feature channels from $256$ to $64$ each. Since the feature maps are $1/4$, $1/8$, $1/16$, $1/32$ and $1/64$ of the original scale, we bilinearly upsample them back to the original size and concatenate them to form a $H_c \times W_c \times 320$ feature tensor. 
To provide the network a notion of where each vertex is, we further append a 2 channel CoordConv layer \cite{coordconv}. The channels represent $x$ and $y$ coordinates with respect to the frame of the crop. Finally, we exploit the bilinear interpolation operation of the spatial transformer network \cite{jaderberg2015spatial} to sample features at the vertex coordinates of the initial polygon from  the feature tensor. 
We denote such $N \times (320+2)$ embedding as $\pmb{z}$. 
\vspace{-1mm}
\paragraph{Deforming network:}
When moving a vertex in a polygon, the two attached edges are subsequently moved as well. The movement of these edges depends on the position of the neighboring vertices. Each vertex thus must be aware of its neighbors and needs a way to communicate with one another in order to reduce unstable and overlapping behavior. 
In this work, we exploit the self-attending Transformer network \cite{transformer} to model such intricate dependencies. We leverage the attention mechanism to propagate the information across vertices and improve the predicted offsets.
More formally, given the vertex embeddings $\pmb{z}$, we first employ three feed-forward neural networks to transform it into $Q(\pmb{z})$, $K(\pmb{z})$, $V(\pmb{z})$, where $Q$, $K$, $V$ stands for Query, Key and Value. We then compute the weightings between vertices by taking a softmax over the dot product $Q(\pmb{z})K(\pmb{z})^T$. Finally, the weightings are multiplied with the keys $V(\pmb{z})$ to propagate these dependencies across all vertices. Such attention mechanism can be written as:
\[
	 Atten(Q(\pmb{z}), K(\pmb{z}), V(\pmb{z}))=%
	 softmax(\frac{Q(\pmb{z})K(\pmb{z})^T}{\sqrt{d_k}})V(\pmb{z}), \label{eq:attn}
\]
where $d_k$ is the dimension of the queries and keys, serving as a scaling factor to prevent extremely small gradients. We repeat the same operation a fixed number of times, 6 in our experiments. After the last Transformer layer, we feed the output to another feed-forward network which predicts $N \times 2$ offsets for the vertices. We add the offsets to the polygon initialization to transform the shape of the polygon.
\subsection{Learning}
We train the deforming network and the feature extraction network in an end-to-end manner.
Specifically, we minimize the weighted sum of two losses. The first penalizes the model for when the vertices deviate from the ground truth. The second regularizes the edges of the polygon to prevent overlap and unstable movement of the vertices. %
\paragraph{Polygon Transforming Loss:}
We make use of the Chamfer Distance loss similar to \cite{HomayounfarMLU18} to move the vertices of our predicted polygon $P$ closer to the ground truth polygon $Q$. The Chamfer Distance loss is defined as:
\[
	L_c(P, Q) = \frac{1}{|P|}\sum_{i}\min_{q \in Q}{ \norm{p_i - q}_2} + \frac{1}{|Q|} \sum_{j} \min_{p \in P}{ \norm{p - q_j}_2} \label{eq:poly_loss}
\]
\vspace{-.5mm}
where $p$ and $q$ are the rasterized edge pixels of the polygons $P$ and $Q$.
To prevent unstable movement of the vertices, we add a  deviation loss on the lengths of the edges $\pmb{e}$ between  vertices. Empirically, we found that without this term the vertices can suddenly shift a large distance, incurring a large loss and causing the gradients to blow up. We define the standard deviation loss as:
$
	L_s(P) = \sqrt{\frac{\sum(\pmb{e} - \bar{\pmb{e}})^2}{n}} \label{eq:std_loss},
	$
where $\bar{\pmb{e}}$ denotes the mean length of the edges.

\section{Experiments}
We evaluate our model in the context of both instance segmentation and interactive annotation settings. 

\vspace{-3.5mm}

\begin{table}[t!]
\footnotesize
\centering
  \begin{tabular}{|c|c|c|cc|cc|}
  \cline{1-7}
  Init & Backbone & \texttt{COCO} & AP  & AP$_{gain}$ &AF&AF$_{gain}$ \\ 
  \hline 
  
  DWT & \texttt{Res101} & - & $18.7$& $+2.2$ &$44.2$ &$+5.8$ \\
  UPSNet & \texttt{Res50} & - &  $33.3$ & $+3.0$  &$41.4$ &$+10.3$\\
  UPSNet & \texttt{Res50} &\checkmark  &  $37.8$ & $+2.4$ &$45.7$ &$+7.8$\\
  UPSNet & \texttt{WRes38+PANet} & \checkmark & $41.4$ & $+1.6$ &$51.1$ &$+4.9$\\
  UPSNet & \texttt{WRes38+PANet+DCN} & \checkmark &  $43.0$ & $+1.6$  &$51.5$ &$+4.2$ \\

  \hline 
    
  \end{tabular}
  \caption{\textbf{Improvement on Cityscapes val instance segmentation initializations:} We report the AP, AF of the initialization and gain in AP, AF from the initialization instances when running our PolyTransform model for Cityscapes val.} %
  \label{tab:improve-cityscapes}
  \vspace{-3mm}
\end{table}

\begin{table}[t!]
\footnotesize
\centering
  \begin{tabular}{|c|c|c|cc|cc|}
  \cline{1-7}
  Init & Backbone & \texttt{COCO} & AP  & AP$_{gain}$ &AF&AF$_{gain}$ \\ 
  \hline 
  
  M-RCNN  & \texttt{Res50} & - &  $28.8$ & $+2.2$ & $44.2$ & $+5.6$\\
  UPSNet& \texttt{Res101} & - &  $31.7$ & $+1.6$& $45.7$ & $+3.2$ \\
  UPSNet & \texttt{Res101} & \checkmark &  $34.2$ & $+1.9$& $45.8$ & $+3.4$\\
  UPSNet & \texttt{WRes38+PANet+DCN} & \checkmark &  $36.1$ & $+1.4$& $50.1$ & $+3.4$\\

  \hline 
    
  \end{tabular}
  \caption{\textbf{Improvement over instance segmentation initializations on the validation of our new self-driving dataset:} We report the AP, AF of the initialization and gain in AP, AF from the initialization instances when running our PolyTransform model for the validation of our new self-driving dataset.}
  \label{tab:improve-uber}
\vspace{-3mm}
\end{table}

\paragraph{Experimental Setup:}
We train our model  on 8 Titan 1080 Ti GPUs using the distributed training framework Horovod \cite{horovod} for 1 day. We use a batch size of 1, ADAM \cite{adam}, 1e-4 learning rate and a 1e-4 weight decay. We augment our data by randomly flipping the images horizontally. During training, we only train with instances whose proposed box has an Intersection over Union (IoU) overlap of over 0.5 with the ground truth (GT) boxes. We train with both instances produced using proposed boxes and GT boxes to further augment the data. For our instance segmentation experiments, we augment the box sizes by $-3\%$ to $+3\%$ during training and test with a $2\%$ box expansion. For our interactive annotation experiments, we train and test on boxes with an expansion of 5px on each side;  we only compute a chamfer loss if the predicted vertex is at least 2px from the ground truth polygon. When placing weights on the losses, we found ensuring the loss values were approximately balanced produced the best result. For our PolyTransform FPN, we use ResNet50 \cite{resnet} as the backbone and use the same pretrained weights from UPSNet \cite{upsnet} on Cityscapes. For our deforming network we do not use pretrained weights.

\begin{figure*} [t!]
\vspace{-0.5cm}
\centering
\setlength\tabcolsep{2pt}
\begin{tabular}{ccc}

\includegraphics[width=.33\textwidth]{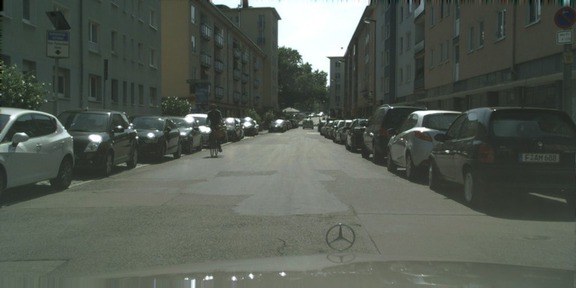} &
\includegraphics[width=.33\textwidth]{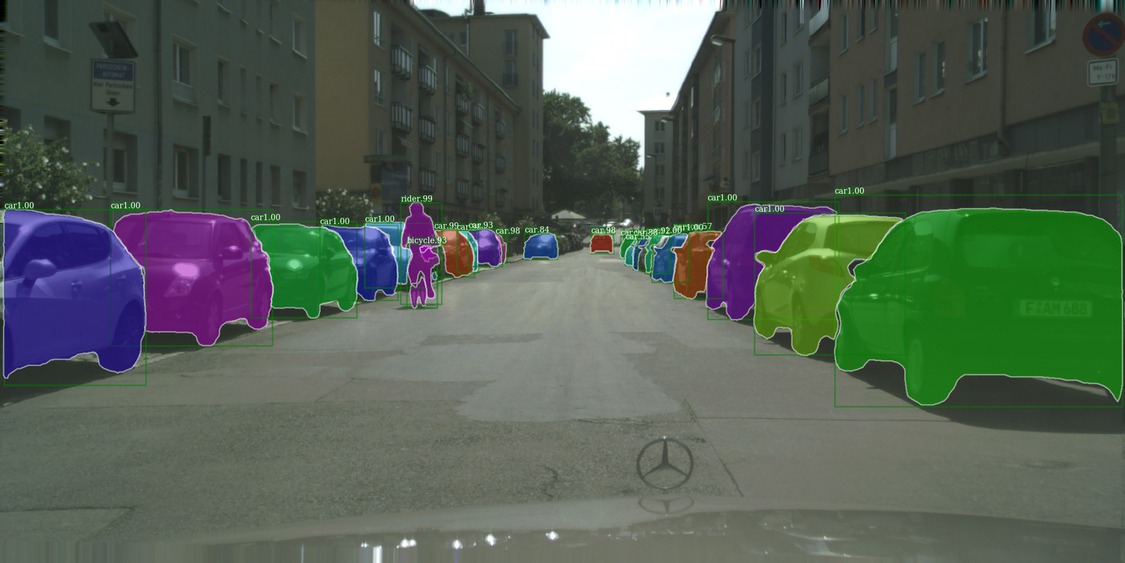} &
\includegraphics[width=.33\textwidth]{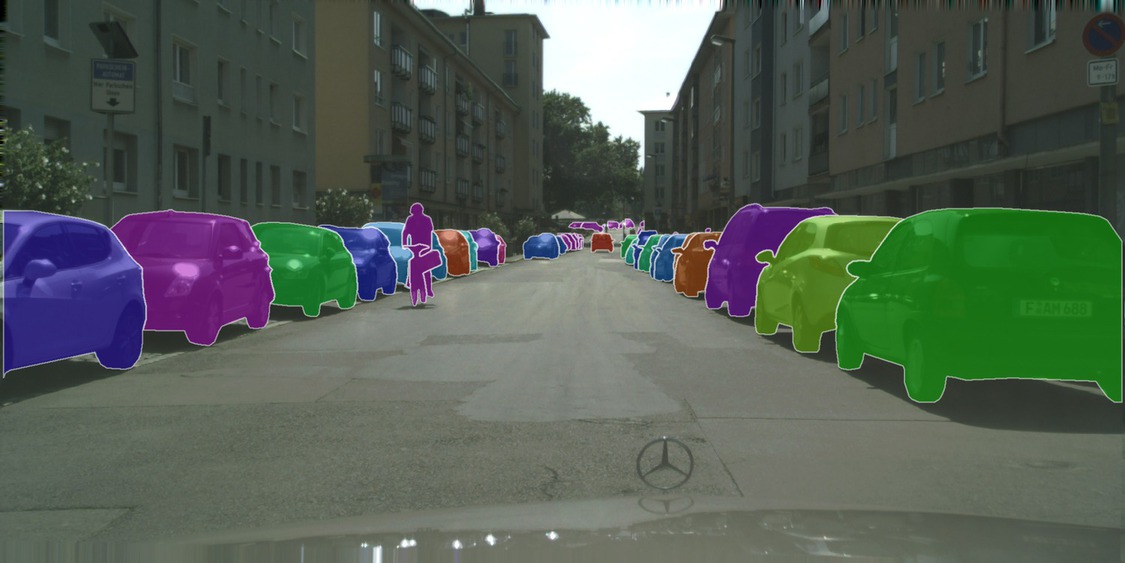}\\

\includegraphics[width=.33\textwidth]{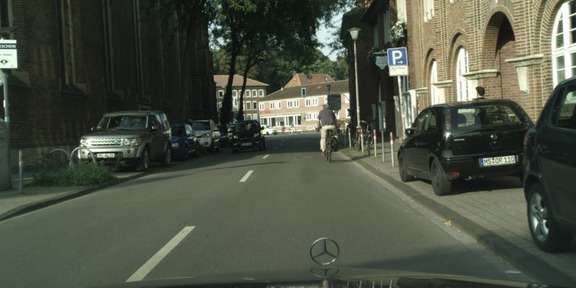} &
\includegraphics[width=.33\textwidth]{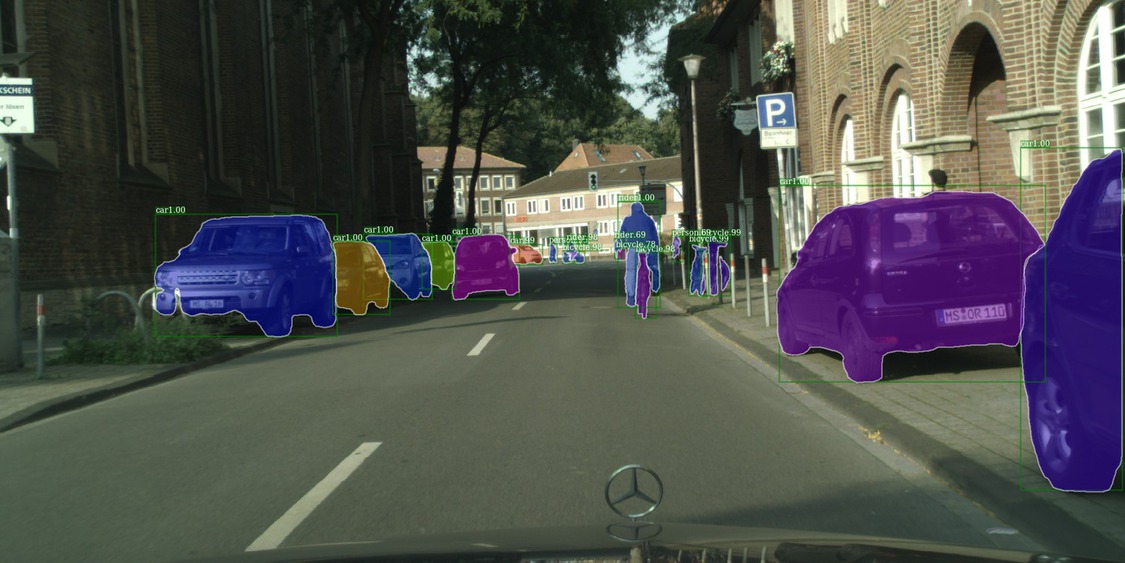} &
\includegraphics[width=.33\textwidth]{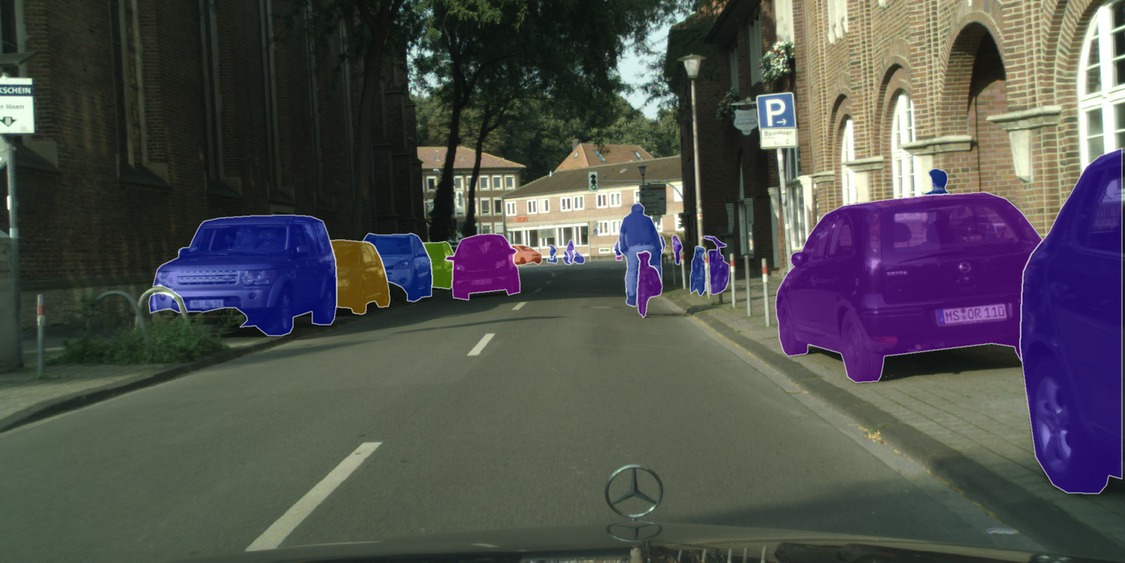}\\

\includegraphics[width=.33\textwidth]{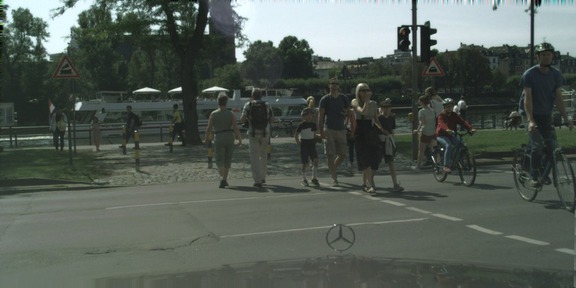} &
\includegraphics[width=.33\textwidth]{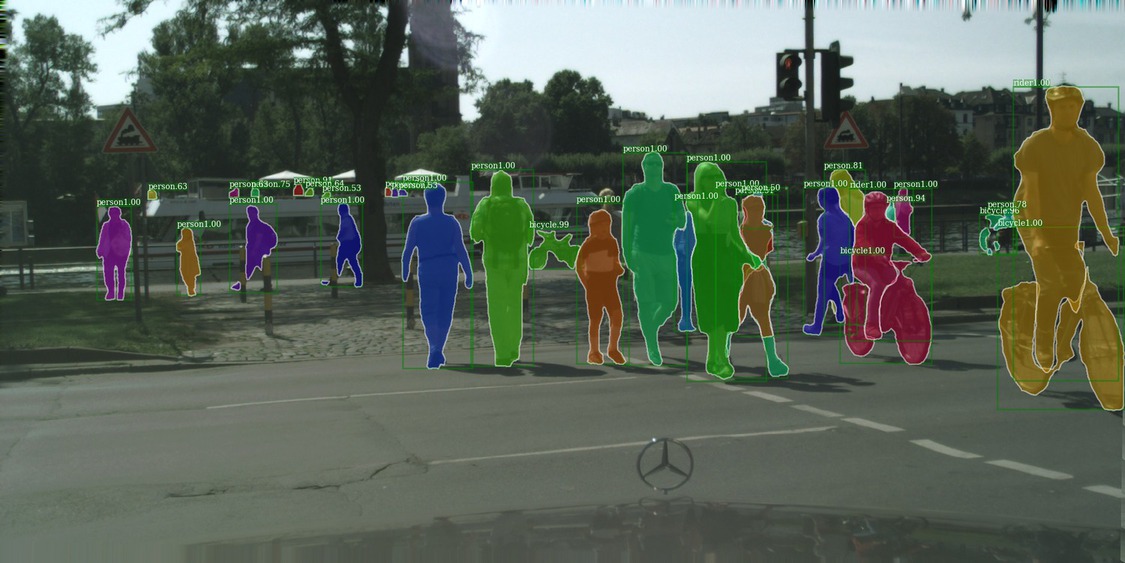} &
\includegraphics[width=.33\textwidth]{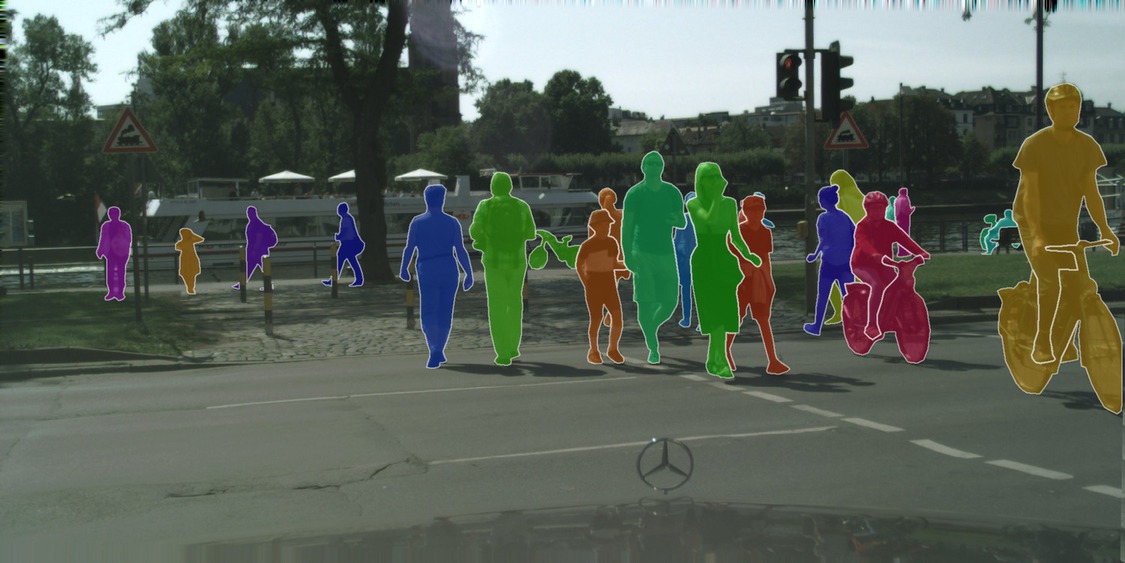}\\

\includegraphics[width=.33\textwidth]{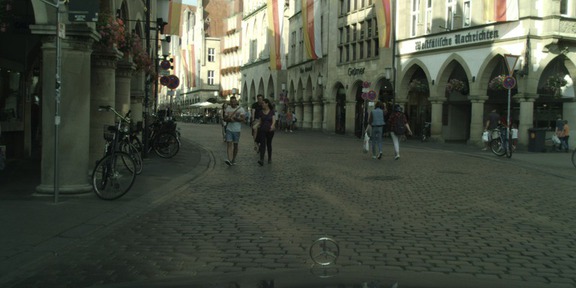} &
\includegraphics[width=.33\textwidth]{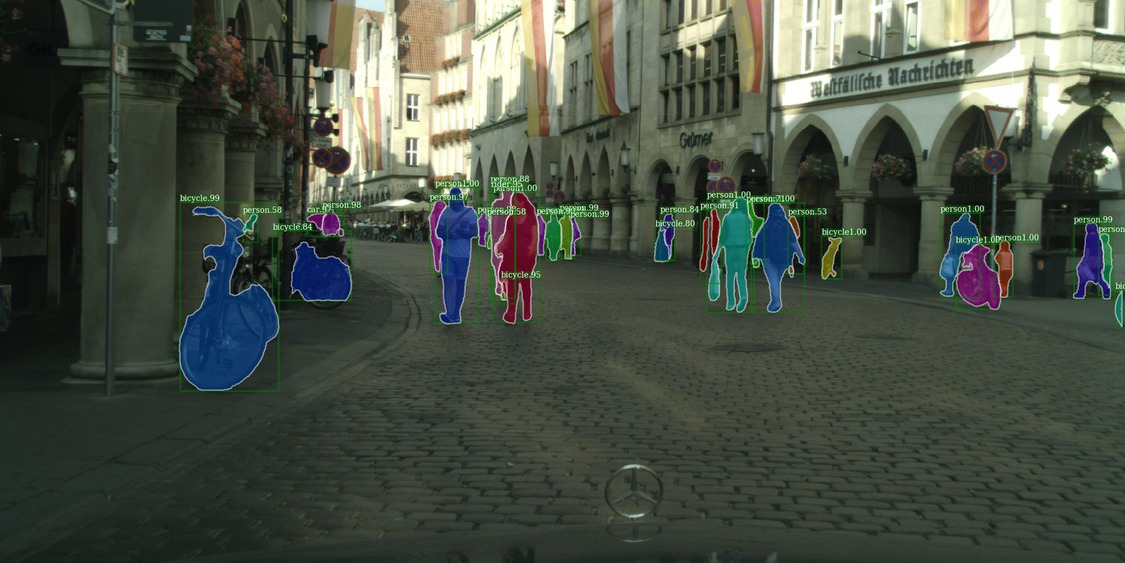} &
\includegraphics[width=.33\textwidth]{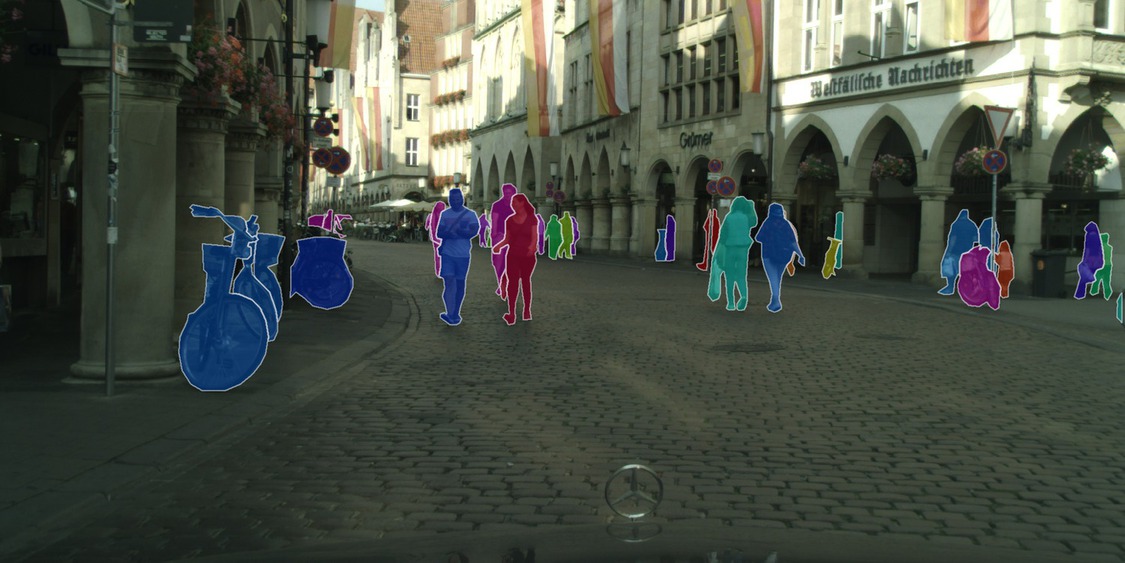}\\

Input Image & Our Instance Segmentation & GT Instance Segmentation

\end{tabular}
\caption{We showcase qualitative instance segmentation results of our model on the Cityscapes validation set.}
\label{fig:results}
\vspace{-4mm}
\end{figure*}

\subsection{Instance Segmentation}  \label{ssec:instance-seg}

\paragraph{Datasets:}
We use Cityscapes \cite{cityscapes} which has high quality pixel-level instance segmentation annotations. The $1024 \times 2048$ images were collected in  27  cities, and  they are split into 2975, 500 and 1525 images for train/val/test. There are 8 instance classes: bicycle, bus, person, train, truck, motorcycle, car and rider.  
We also report results on a new  dataset we collected. It consists of 10235/1139/1186  images for train/val/test split annotated with 10 classes: car, truck, bus, train , person, rider, bicycle with rider, bicycle, motorcycle with rider and motorcycle. Each image is of size $1200 \times 1920$. %

\begin{table}[t!]
\centering
{\begin{footnotesize}
  \begin{tabular}{|c|cc|}
  \hline
  Cityscapes (\texttt{fine+COCO}) & AP  & AF  \\ 
  \hline 
  UPSNet &$43.0$ &$51.5$\\
  Baseline 1
  &$43.8$ &$52.6$\\
  Baseline 2
  &$43.5$ &$52.4$\\
  Ours &$\bf{44.6}$ &$\bf{55.7}$ \\   
  \hline 
  \end{tabular}
  \end{footnotesize}}
  \caption{\textbf{
Comparison with naive refiners on Cityscapes val set.} }
  \label{tab:inst-val-results}
  \vspace{-4mm}
\end{table}

\vspace{-2mm}

\paragraph{Metrics:}
For our instance segmentation results, we report the average precision (AP and AP$_{50}$) for the predicted mask. Here, the AP is computed at 10 IoU overlap thresholds ranging from 0.5 to 0.95 in steps of 0.05 following \cite{cityscapes}. AP$_{50}$ is the AP at an overlap of 50\%. 
We also introduce a new metric that focusses on  boundaries.  In particular, we use a metric similar to \cite{wang2019delse, Perazzi2016} where a precision, recall and F1 score is computed for each mask,  where the prediction is correct if it is within a certain distance threshold from the ground truth. We use a threshold of 1px, and only compute the metric for  true positives. We   use the same 10 IoU overlap thresholds ranging from 0.5 to 0.95 in steps of 0.05 to determine the true positives. Once we compute the F1 score for all classes and thresholds, we take the average over all  examples to get AF. %

\vspace{-2mm}

\paragraph{Instance Initialization:}
We want to use a strong instance initialization to show that we can still improve the results.  We take the publicly available UPSNet \cite{upsnet}, and replace its backbone with WideResNet38 \cite{wideresnet} and add all the elements of PANet \cite{panet} except for the synchronized batch normalization (we use group normalization instead). We then pretrain on COCO  and use deformable convolution (DCN) \cite{dcn} in the backbone. 

\paragraph{Comparison to SOTA:}
As shown in \ref{tab:inst-test-results} we outperform all baselines in every metric  on the val and test sets of Cityscapes.  
We achieve a new state-of-the-art test result of 40.1AP. This  outperforms PANet by 3.7 and 2.8 points in AP and AP$_{50m}$ respectively. It also ranks number 1 on the official Cityscapes leaderboard. 
We report the results on our new dataset in Table \ref{tab:inst-test-results-uber}. We achieve the strongest test AP result in this leaderboard. We see that we improve over PANet by 6.2 points in AP and UPSNet by 3.8 points in AP. 

\vspace{-2mm}

\paragraph{Robustness to Initialization:}
We  report the improvement over different instance segmentation networks used as initialization in Table \ref{tab:improve-cityscapes} on Cityscapes, showing  significant and consistent improvements in val AP across all models. When we train our model on top of the DWT \cite{bai2017deep} instances we see an improvement of $+2.2$, $+5.8$ points in AP and AF. We also train on top of the UPSNet results from the original paper along with UPSNet with \texttt{WRes38+PANet} as a way to reproduce the current SOTA val AP of PANet. We show an improvement of $+1.6$, $+4.9$ points in AP and AF. Finally we  improve on our best initialization by $+1.6, +4.2$ AP points in AP and AF. As we can see, our boundary metric sees a very consistent $4\%-10\%$ gain in AF across all  models. This suggests that our approach  significantly improvs the instances at the boundary. We notice that a large gain in AP (\texttt{WRes38+PANet} to \texttt{WRes38+PANet+DCN}) does not necessarily translate to a large gain in AF, however, our model will always provide a significant increase in this metric.
We also report the validation AP improvement over different instance segmentation networks in Table \ref{tab:improve-uber} for our new dataset. We see that we can improve on Mask R-CNN \cite{mask-rcnn} by $+2.2$, $+5.6$ points in AP, AF. For the different UPSNet models, we improve upon it between 1.4-2.2 AP points. Once again, our model shows a consistent and strong improvement over all initializations. We also see a very consistent $3\%-6\%$ gain in AF across all the models. %

\begin{figure*} [t!]
\vspace{-0.5cm}
\centering
\setlength\tabcolsep{1pt}
\begin{tabular}{cccc}

\raisebox{15px}{\rotatebox{90}{\small Input Image}}
\includegraphics[width=.245\textwidth]{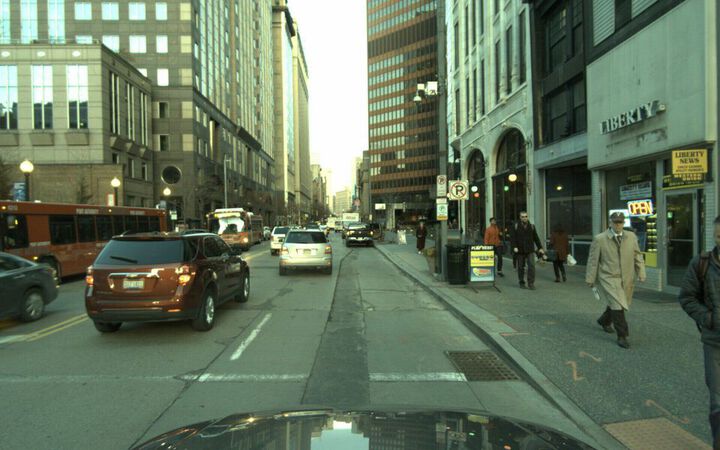} &
\includegraphics[width=.245\textwidth]{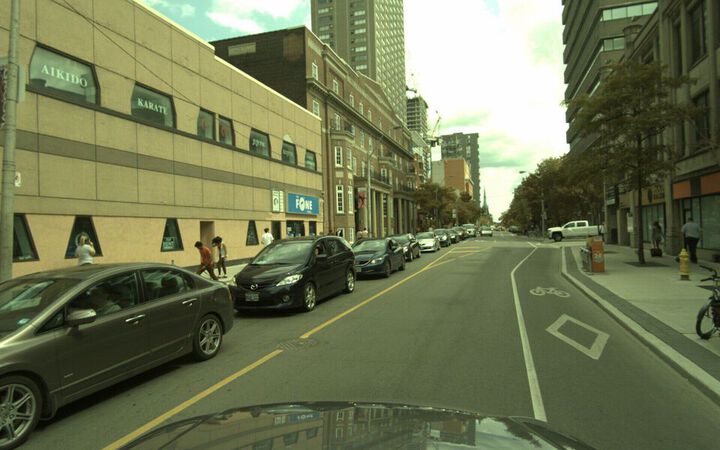}
&
\includegraphics[width=.245\textwidth]{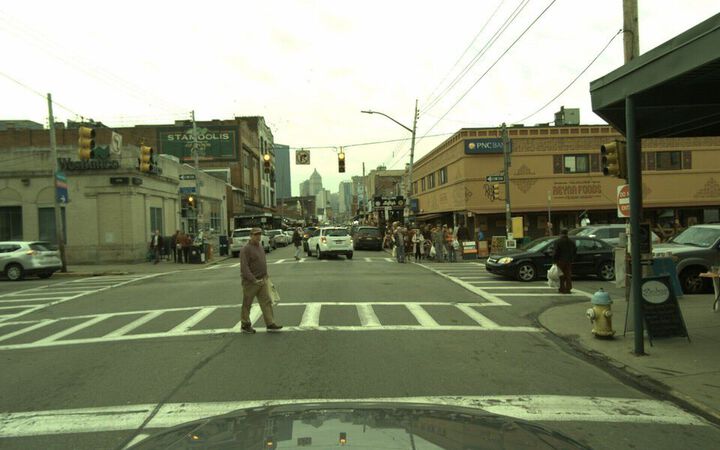}
&
\includegraphics[width=.245\textwidth]{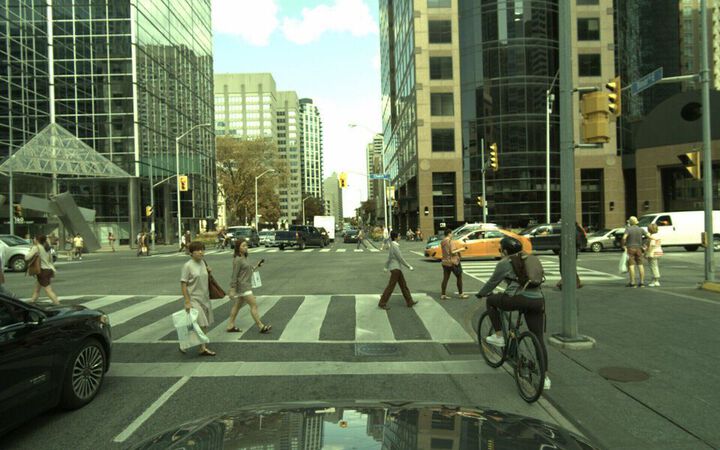}
\\

\raisebox{8px}{\rotatebox{90}{\small Our Instance Seg}}
\includegraphics[width=.245\textwidth]{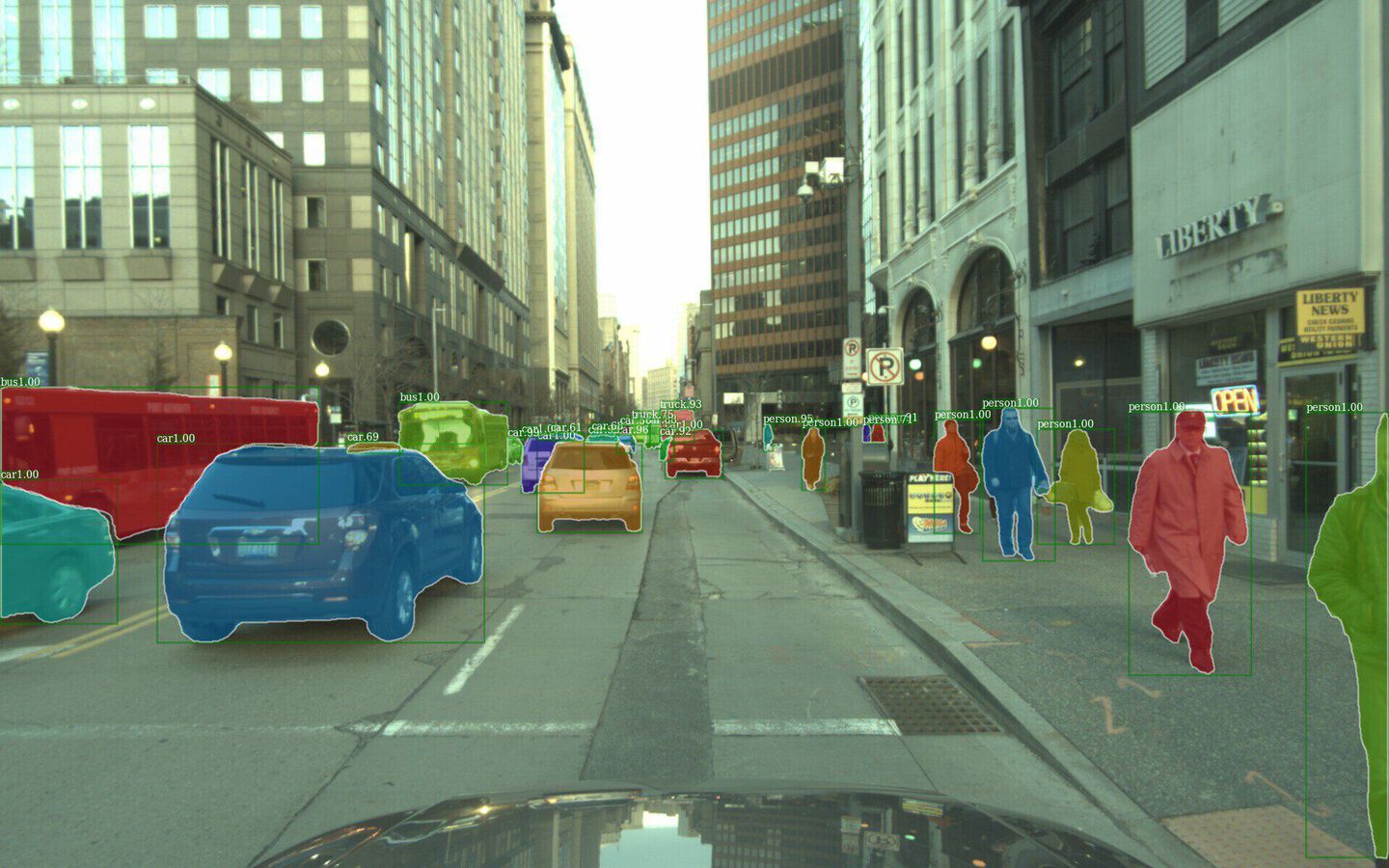}  &
\includegraphics[width=.245\textwidth]{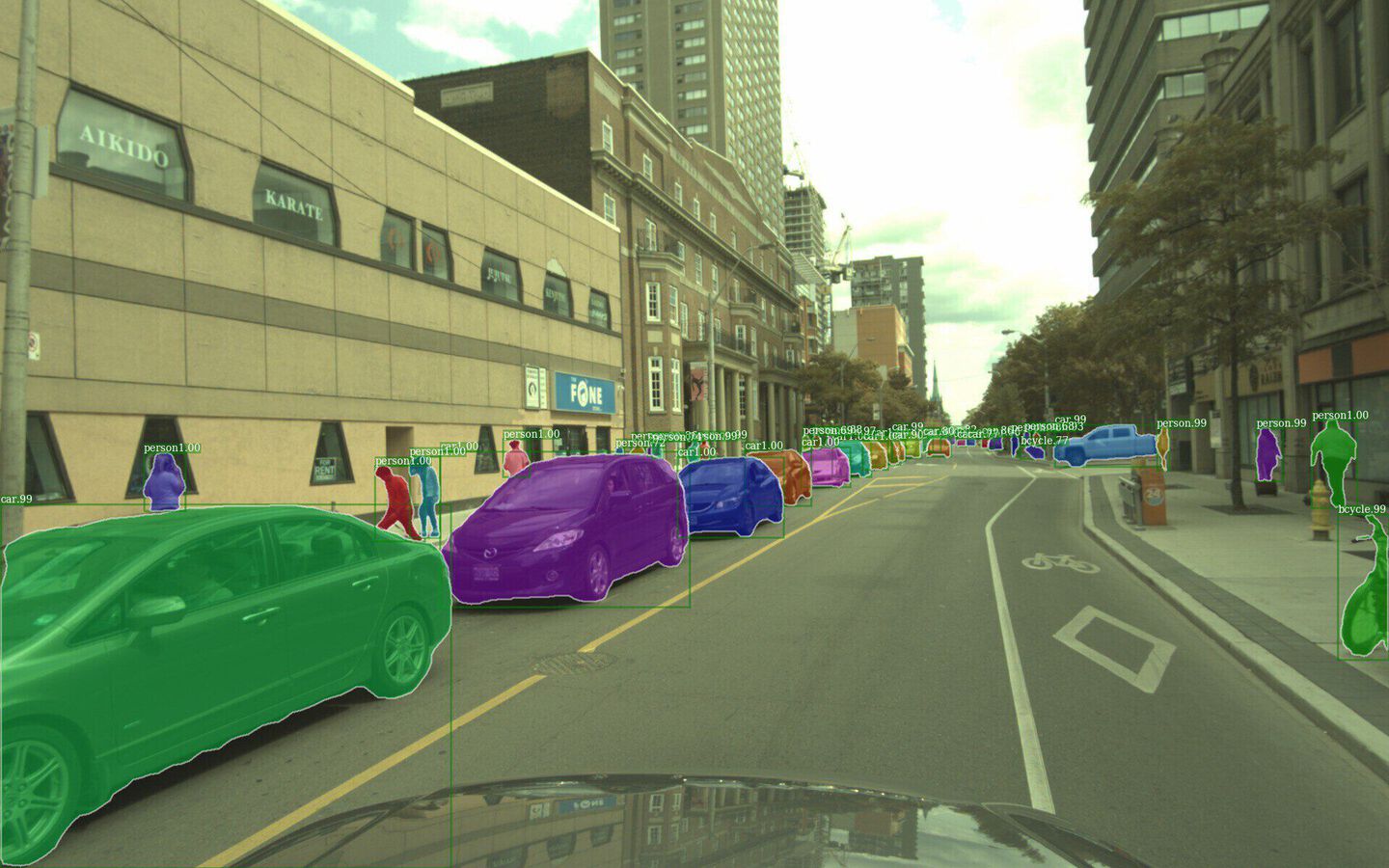}
&
\includegraphics[width=.245\textwidth]{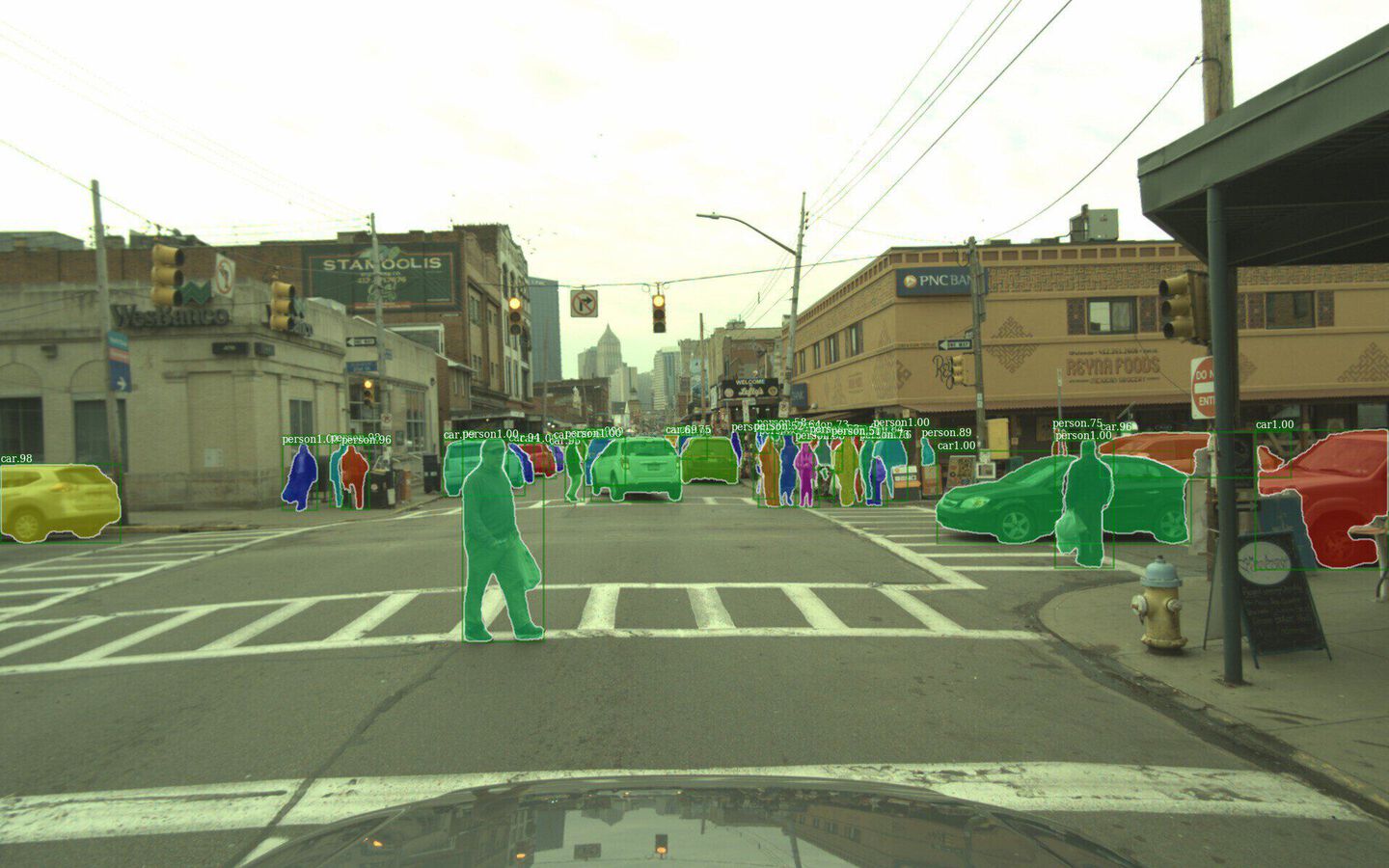}
&
\includegraphics[width=.245\textwidth]{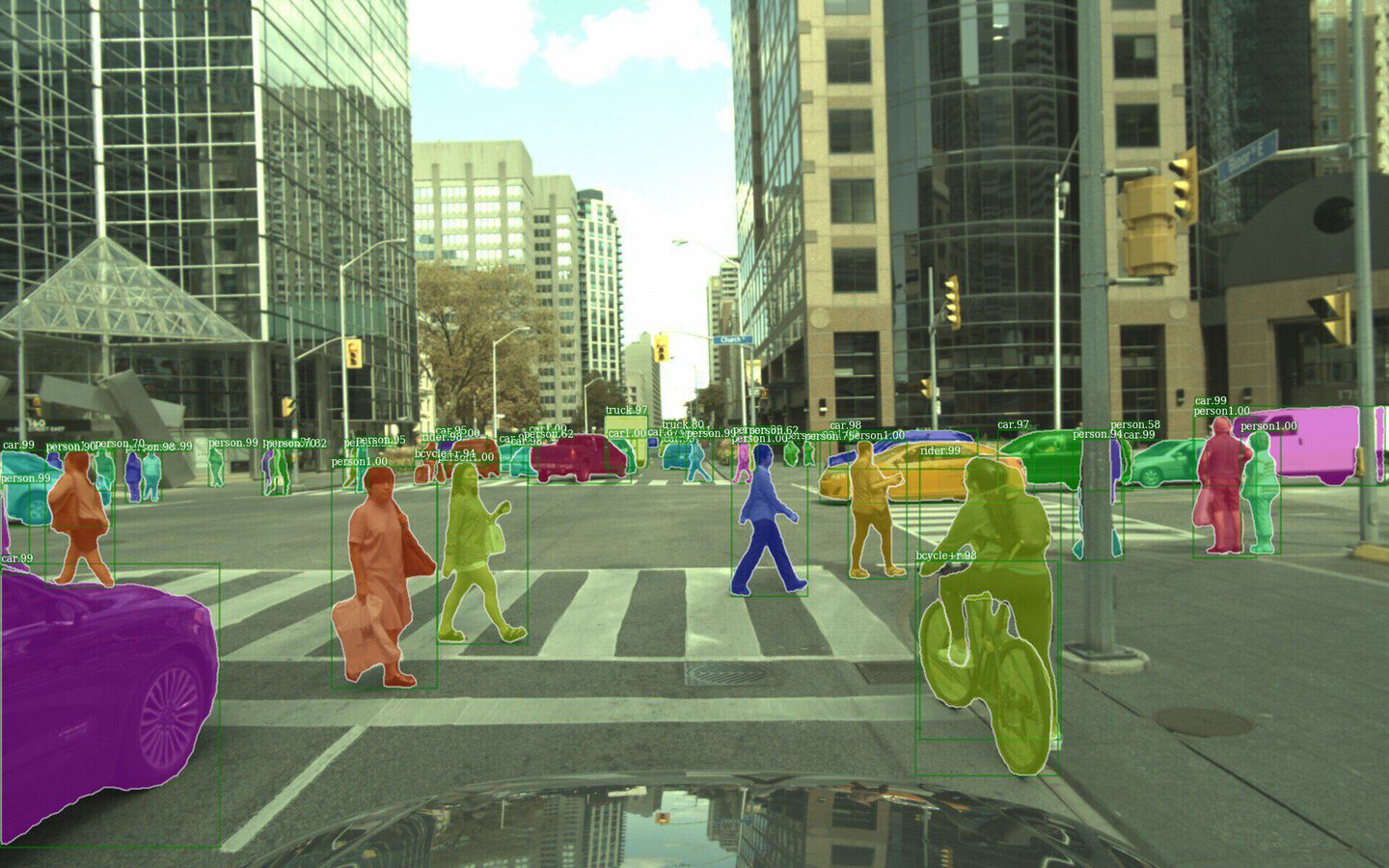}
\\

\raisebox{8px}{\rotatebox{90}{\small GT Instance Seg}}
\includegraphics[width=.245\textwidth]{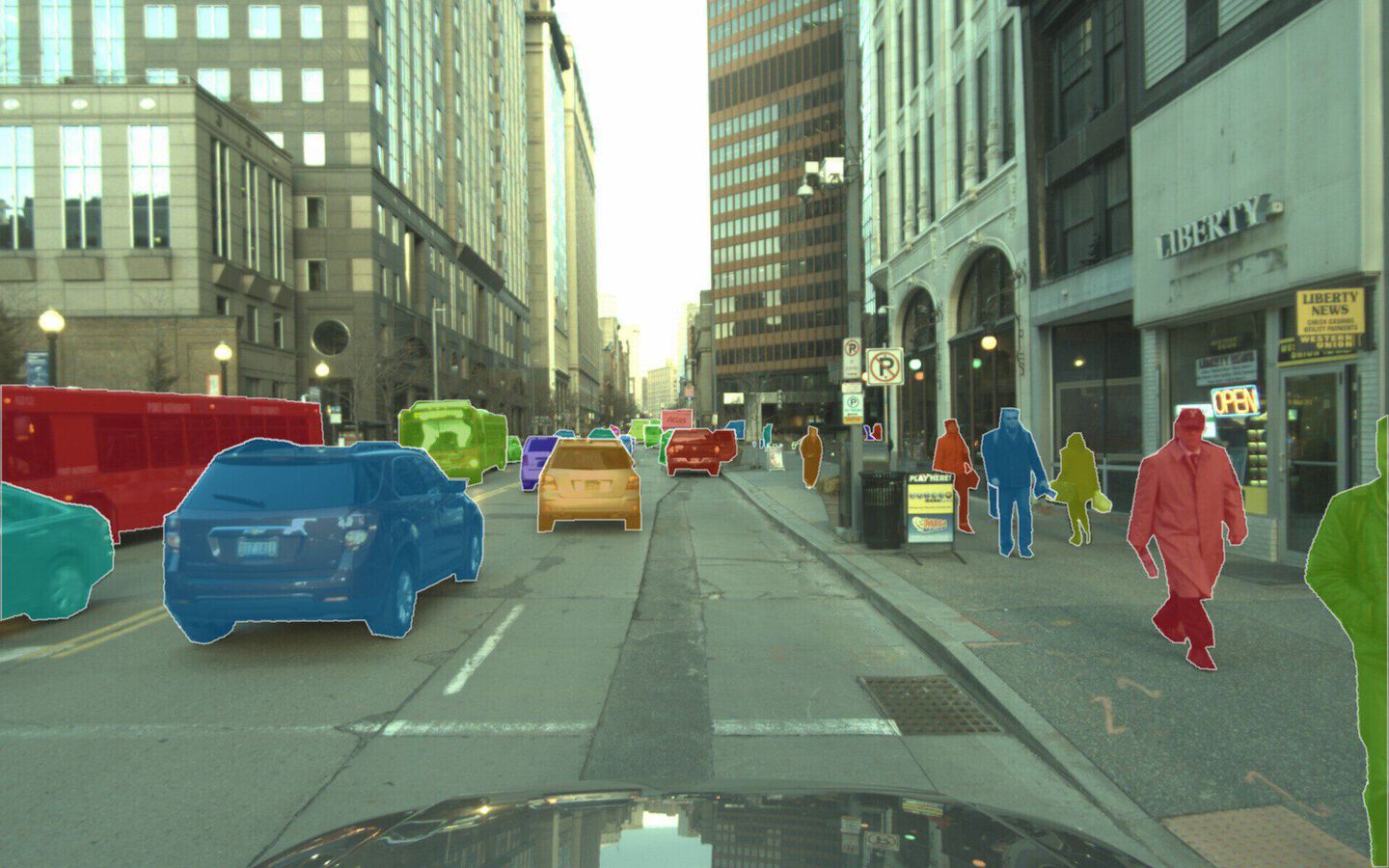}
&
\includegraphics[width=.245\textwidth]{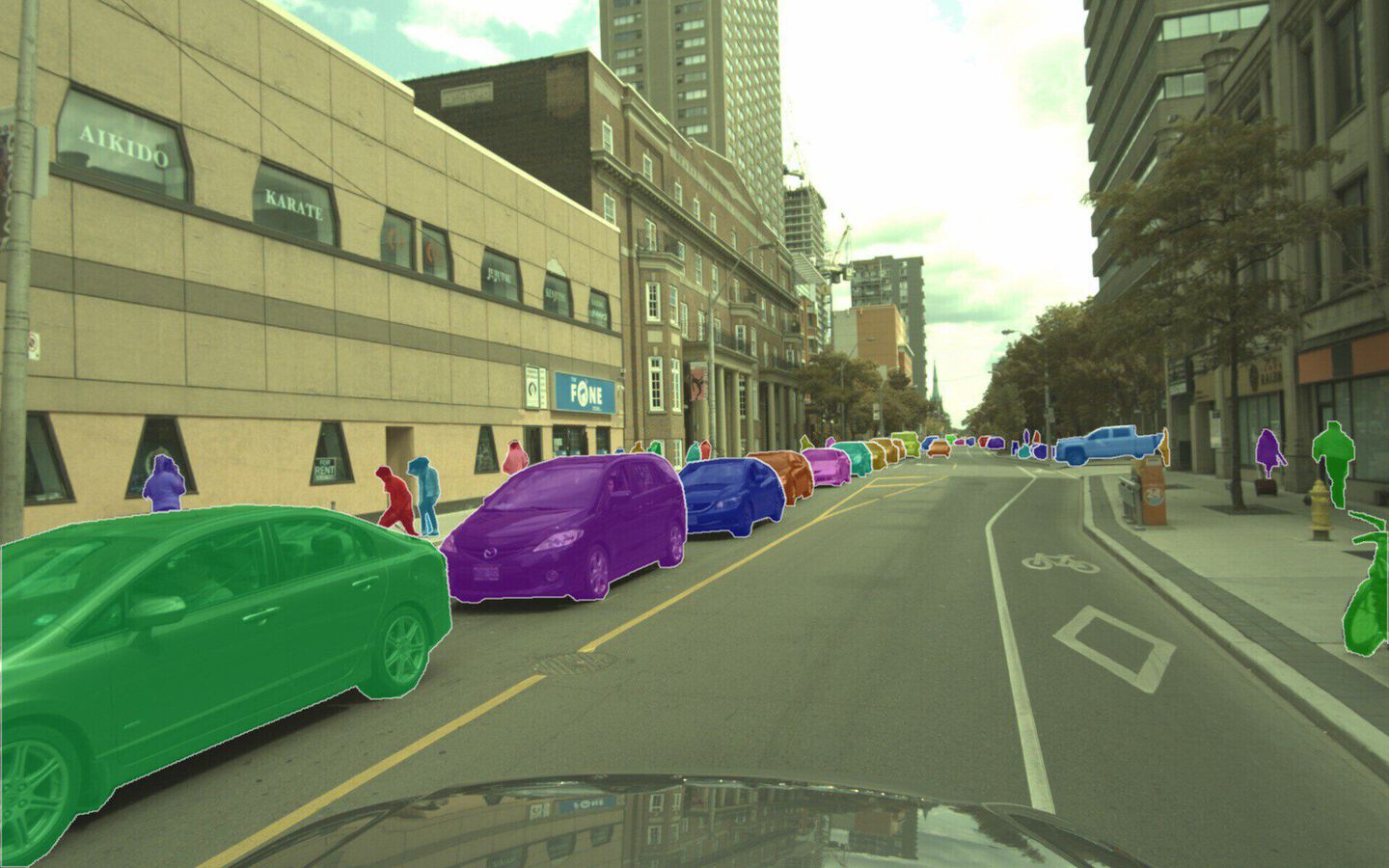}
  &
\includegraphics[width=.245\textwidth]{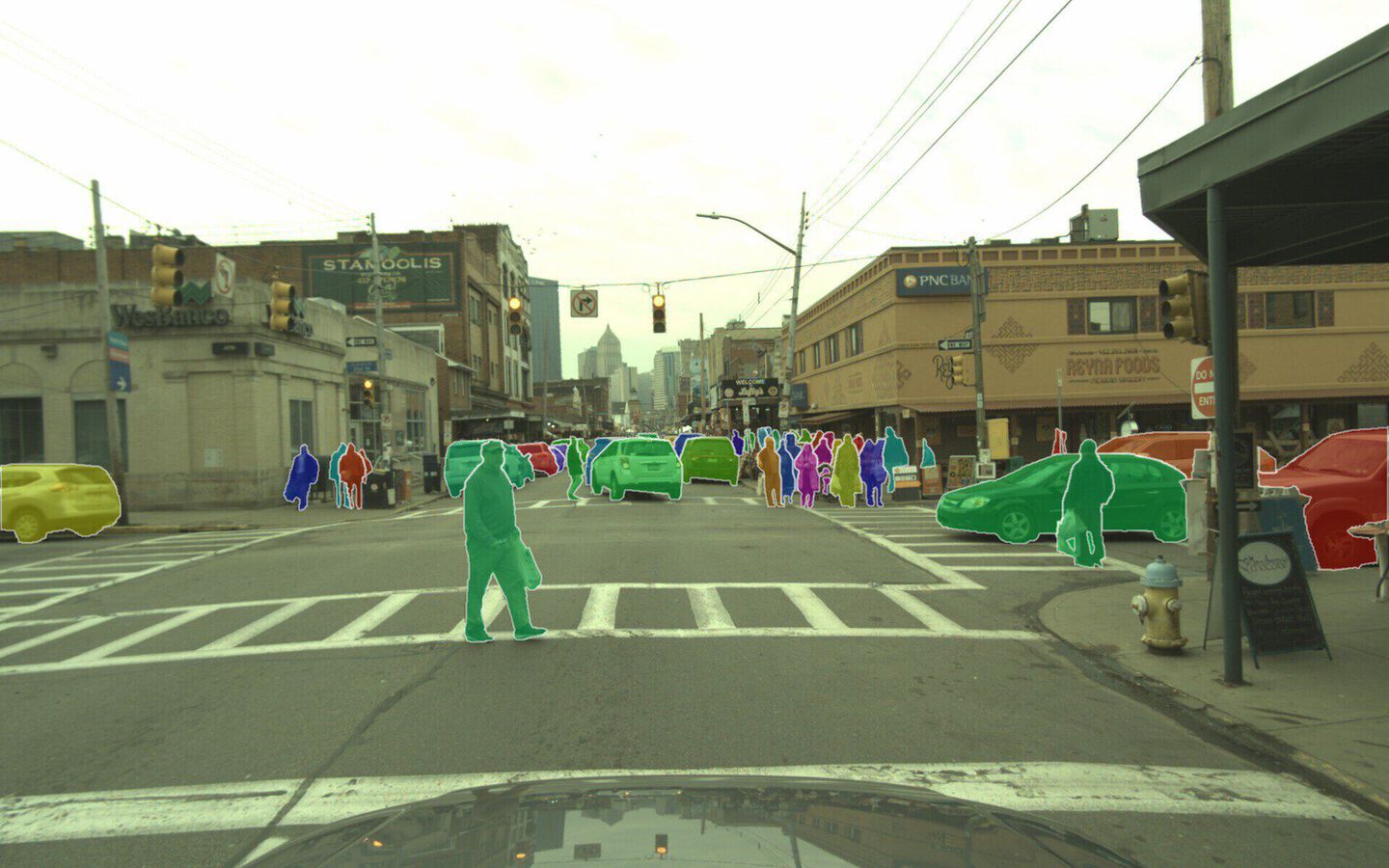}  &
\includegraphics[width=.245\textwidth]{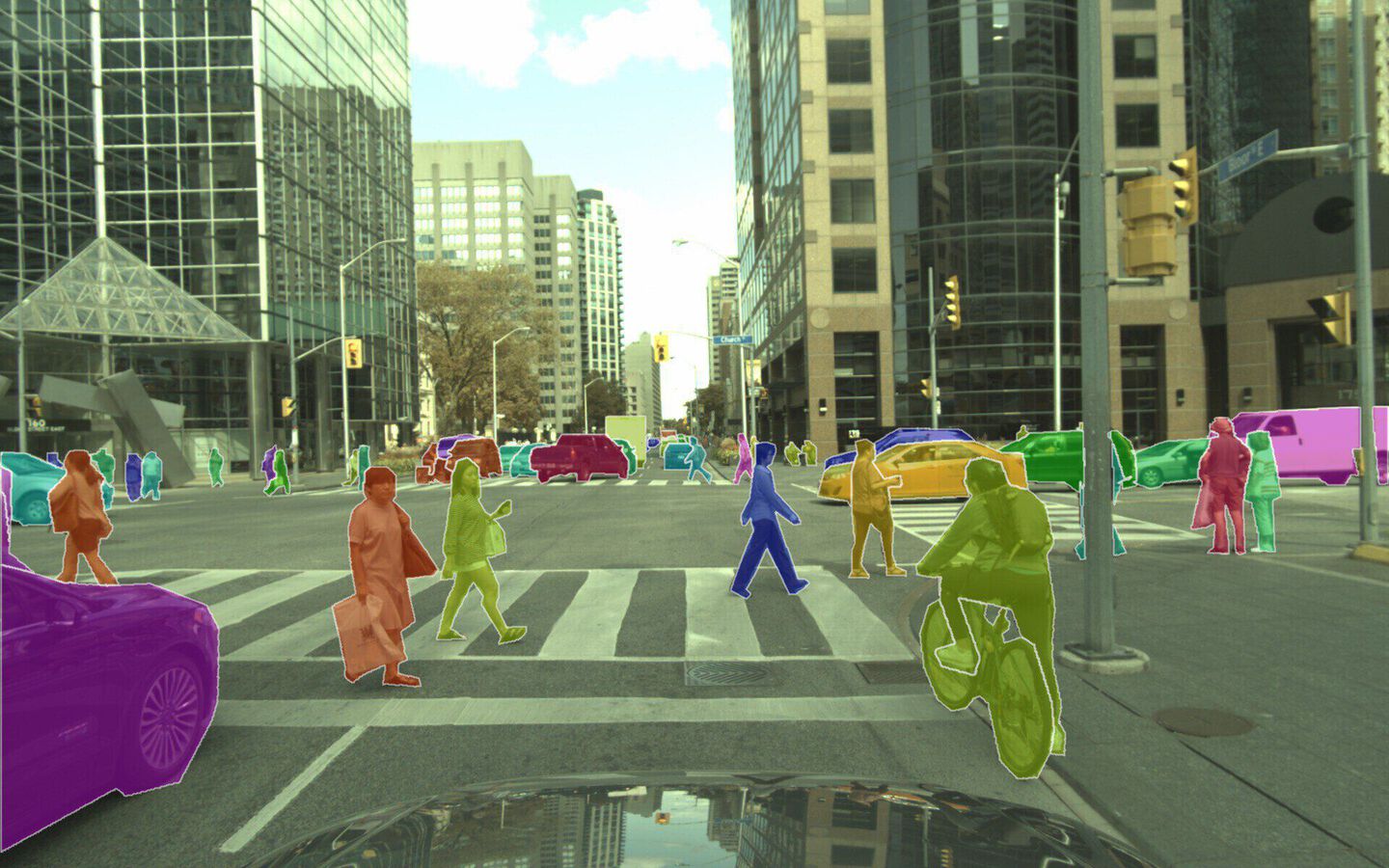}\\

\end{tabular}
\caption{We showcase the qualitative instance segmentation results of our model on the validation set of our new self-driving dataset}
\label{fig:results-uber}
\end{figure*}

\begin{table*}[t!]
\centering
  \begin{tabular}{|c|c|cccccccc|cc|}
  \cline{2-12}
  \multicolumn{1}{c|}{}  & Mean & \multicolumn{1}{c}{bicycle} & \multicolumn{1}{c}{bus} & \multicolumn{1}{c}{person} & \multicolumn{1}{c}{train} & \multicolumn{1}{c}{truck} & \multicolumn{1}{c}{mcycle} & \multicolumn{1}{c}{car} & rider & \multicolumn{1}{c}{F$_{1px}$} & \multicolumn{1}{c|}{F$_{2px}$}  \\ 
  \hline 
  DEXTR* \cite{dextr} 
    &$79.11$ &$71.92$  &$87.42$ &$78.36$ &$78.11$ &$84.88$ &$72.41$ &$84.62$ &$75.18$ &$54.00$ &$68.60$ \\ 
  Deep Level Sets \cite{wang2019delse} 
    &$80.86$ &$\bf{74.32}$ &$\bf{88.85} $ &$80.14$ &$\bf{80.35}$ &$ 86.05$ &$ 74.10 $ &$86.35$ &$  76.74$ &$60.29$ &$74.40$ \\
  Ours 
    &$\bf{80.90}$ &$74.22$ &$88.78$ &$\bf{80.73}$ &$77.91$ &$\bf{86.45}$ &$\bf{74.42}$  &$\bf{86.82}$ &$\bf{77.85}$ &$\bf{62.33}$ &$\bf{76.55}$\\
  \hline 

  \end{tabular}
  \caption{\textbf{Interactive Annotation (Cityscapes Stretch):} This table shows our IoU \% performance in the setting of annotation where we are given the ground truth boxes. DEXTR* represents DEXTR without extreme points.}
  \label{tab:anno-results-stretch}
\end{table*}

\begin{table*}[t!]
\centering
  \begin{tabular}{|c|c|cccccccc|cc|}
  \cline{2-12}
  \multicolumn{1}{c|}{}  & Mean & \multicolumn{1}{c}{bicycle} & \multicolumn{1}{c}{bus} & \multicolumn{1}{c}{person} & \multicolumn{1}{c}{train} & \multicolumn{1}{c}{truck} & \multicolumn{1}{c}{mcycle} & \multicolumn{1}{c}{car} & rider & \multicolumn{1}{c}{F$_{1px}$} & \multicolumn{1}{c|}{F$_{2px}$}  \\ 
  \hline 
    
  Polygon-RNN \cite{polygon-rnn} 
    &$61.40$ &$52.13$ &$69.53$ &$63.94$ &$53.74$ &$68.03$ &$52.07$ &$71.17$ &$60.58$ &$-$ &$-$ \\ 
  Polygon-RNN++  \cite{polygon-rnn++} 
    &$71.38$ &$63.06$ &$81.38$ &$72.41$ &$64.28$ &$78.90$ &$62.01$ &$79.08$ &$69.95$ &$46.57$ &$62.26$ \\
  Curve GCN  \cite{ling2019fast} 
    &$73.70$ &$67.36$ &$ 85.43 $ &$73.72$ &$ 64.40$ &$ 80.22$ &$ 64.86 $ &$81.88$ &$ 71.73$ &$47.72$ &$63.64$\\
  Deep Level Sets \cite{wang2019delse}
    &$73.84$ &$67.15$ &$83.38$ &$73.07$ &$69.10$ &$80.74$ &$65.29 $ &$81.08$ &$ 70.86 $ &$48.59$ &$64.45$\\
  Ours 
    &$\bf{78.76}$ &$\bf{72.97}$ &$\bf{87.53}$ &$\bf{78.58}$ &$\bf{72.25}$ &$\bf{85.08}$ &$\bf{72.50}$  &$\bf{85.36}$ &$\bf{75.83}$ &$\bf{56.89}$ &$\bf{71.60}$\\
  \hline 
  
  \end{tabular}
  \caption{\textbf{Interactive Annotation (Cityscapes Hard):} This table shows our IoU \% performance in the setting of annotation where we are given the ground truth boxes.}
  \label{tab:anno-results-hard}
  \vspace{-5mm}
\end{table*}

\vspace{-2mm}

\paragraph{Annotation Efficiency:}
We conduct an experiment where we ask crowd-sourced labelers to annotate 150 images from our new dataset with instances larger than 24x24px for vehicles and 12x14px for pedestrians/riders. We performed a control experiment where the instances are annotated completely from scratch (without our method) and a parallel experiment where we use our model to output the instances for them to fix to produce the final annotations. In the fully manual experiment, it took on average 60.3 minutes to annotate each image. When the annotators were given the PolyTransform output to annotate on top of, it took on average 39.4 minutes to annotate each image. Thus reducing  35\% of the time required to annotate the images. This resulted in significant cost savings. %

\vspace{-2mm}

\paragraph{Naive refiner:} We implemented two  baselines that apply a semantic segmentation network on top of the initial mask. 
1) We replace PolyTransform with a refinement network inspired by DeepLabV3 \cite{deeplabv3} and PWC-Net \cite{Sun2018PWC-Net} . 
It takes as input the same initialization mask, the cropped RGB image and the cropped feature, and exploits a series of convolutions to refine the binary mask. 
2) We add an extra head to  UPSNet, with the initialization mask and the cropped feature as input to refine the binary mask. The head's architecture  is similar to that of the semantic head (i.e., uses the features from  UPSNet's FPN).
For fairness, the number of parameters of both baselines are similar to PolyTransform. 
As shown in Tab. \ref{tab:inst-val-results}, our approach  performs the best.

\paragraph{Timing:}
Our model takes 575 ms to process each image on Cityscapes. This can easily be improved with more GPU memory, as this will allow to batch all the instances. Furthermore, the hidden dimension of the FPN can be tuned to speed up the model.

\vspace{-1mm}

\paragraph{Qualitative Results:}
We show qualitative results of our model on the validation set in Figure \ref{fig:results}. In our instance segmentation outputs we see that in many cases our model is able to handle occlusion. For example, in row 3, we see that our model is able to capture the feet of the purple and blue pedestrians despite their feet being occluded from the body. %
We also show qualitative results on our new dataset in Figure \ref{fig:results-uber}. We see that our model is able to capture precise boundaries, allowing it to capture difficult shapes such as car mirrors and pedestrians.

\setlength{\tabcolsep}{1pt}
\begin{table}[t!]
\footnotesize
\centering
  \begin{tabular}{|c|c|c|cc|cc|cc|}
  \cline{2-9}
  \multicolumn{1}{c|}{}  & BBone & \texttt{COCO} & mIOU &mIOU$_{gain}$ & F$_{1}$ & F$_{1,gain}$ & F$_{2}$ & F$_{2,gain}$  \\ 
  \hline 
  
  FCN & \texttt{R50} & - & $79.93$ & $+0.15$ & $59.43$ & $+1.53$ & $73.64$ & $+1.30
$ \\
  FCN & \texttt{R101}  & - & $80.94$ & $+0.11$ & $60.64$ & $+1.14$ & $74.78$ & $+1.06$ \\
  FCN & \texttt{R101}& \checkmark & $80.65$ & $+0.08$ & $59.21$ & $+1.39$ & $73.47$ & $+1.10$ \\
  DeepLabV3 & \texttt{R50} & - & $80.41$ & $+0.17$ & $59.70$ & $+1.51$ & $73.81$ & $+1.48$ \\
  DeepLabV3  & \texttt{R101}& - & $80.93$ & $+0.09$ & $60.50$ & $+1.18$ & $74.44$ & $+1.33$ \\
  DeepLabV3+ & \texttt{R101} & \checkmark & $80.90$ & $+0.08$ & $61.10$ & $+1.23$ & $75.25$ & $+1.30$ \\

  \hline 
    
  \end{tabular}
  \caption{\textbf{Improvement on Cityscapes Stretch segmentation initializations:} We report the metric improvements when running our PolyTransform model on different models. We report our model results trained on FCN \cite{fcn} and DeepLabV3 \cite{deeplabv3}. DeepLabV3+ uses the class balancing loss from \cite{dextr}. We report on models with various backbones (Res50 vs Res101) and also with and without pretraining on COCO \cite{coco}.}
  \label{tab:improve-annot}
\vspace{-5mm}
\end{table}

\begin{figure} [!t]
\centering
\setlength\tabcolsep{2pt}
\begin{tabular}{ccc}
\includegraphics[width=.2\textwidth]{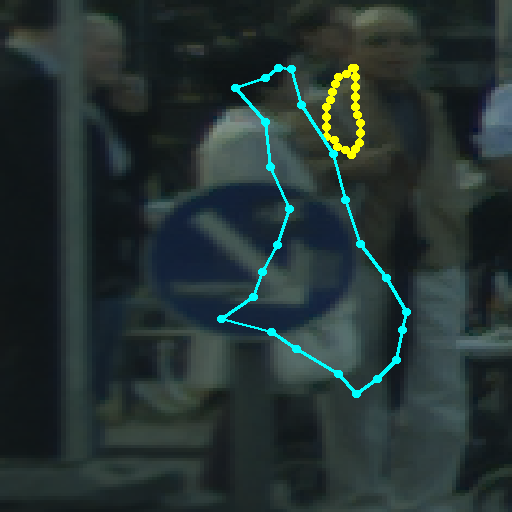} &
\includegraphics[width=.2\textwidth]{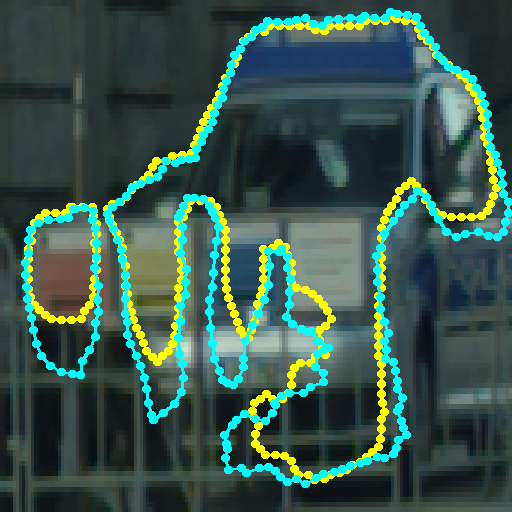} &

\end{tabular}
\caption{\textbf{Failure modes:}  (Left) Our model fails because the initialization is poor. (Right) The model fails because of complex occlusion. (Yellow: Initialization; Cyan: Ours)}
\vspace{-3mm}
\label{fig:failure}
\end{figure}

\paragraph{Failure Modes:} 
Our model can  fail when the initialization is poor (left image in Figure \ref{fig:failure}). Despite being able to handle occlusion, our model can still fail when the occlusion is complex or ambiguous as seen in the right of Figure \ref{fig:failure}. Here there is a semi-transparent fence blocking the car.

\subsection{Interactive Annotation} \label{ssec:annot}
The goal is to annotate an object with a polygon given its ground truth bounding box. The idea is that the annotator provides a ground truth box and our model works on top of it to output a polygon representation of the object instance.

\vspace{-.4mm}

\paragraph{Dataset:}
We follow \cite{polygon-rnn} and split the Cityscapes dataset such that the original val set is the test set and two cities from the training (Weimar and Zurich) form the val set. \cite{wang2019delse} further splits this dataset into two settings: 1) Cityscapes Hard, where the ground truth bounding box is enlarged to form a square and then the image is cropped. 2) Cityscapes Stretch, where the ground truth bounding box along with the image is stretched to a square and then cropped.

\vspace{-.4mm}

\paragraph{Metric:}
To evaluate our model for this task, we report the intersection over union (IoU) on a per-instance basis and average for each class. Then, following \cite{polygon-rnn} this is averaged across all classes. We also report the  boundary metric reported in \cite{wang2019delse, Perazzi2016}, which computes the F measure along the contour for a given threshold. The thresholds used are 1 and 2 pixels as Cityscapes contains a lot of small instances.

\vspace{-.4mm}

\paragraph{Instance Initialization:}
For our best model we use a variation of DeepLabV3 \cite{deeplabv3}, which we call DeepLabV3+ as the instance initialization network. The difference is that we train DeepLabV3 with the class balancing loss used in \cite{dextr}.

\vspace{-.4mm}

\paragraph{Comparison to SOTA:}
Tables \ref{tab:anno-results-stretch} and \ref{tab:anno-results-hard} show  results  on the test set in both Cityscapes Stretch and Hard. For Cityscapes Stretch, we see that our model significantly outperforms the SOTA in the boundary metric, improving it by up to 2\%. Unlike the Deep Level Sets \cite{wang2019delse} method which outputs a pixel wise mask, our method outputs a polygon which allows for it to be amenable to modification by an annotator by simply moving the vertices. For Cityscapes Hard, our model outperforms the SOTA by 4.9\%, 8.3\% and 7.2\% in mean IOU, F at 1px and F at 2px respectively. 

\paragraph{Robustness to Initalization:}
We also report  improvements over different segmentation initializations  in Table \ref{tab:improve-annot}, the results are on the test set. Our models are trained on various backbone initialization models (FCN \cite{fcn} and DeepLabV3  \cite{deeplabv3} with and without pretraining on COCO \cite{coco}). Our model is able to consistently and significantly improve the boundary metrics at 1 and 2 pixels by up to 1.5\% and we improve the  IOU between 0.1-0.2\%. We also note that the difference in mean IOU between FCN and DeepLabV3 is very small (at most 0.5\%) despite DeepLabV3 being a much stronger segmentation model. We argue that the margin for mean IOU improvement is very small for this dataset. 

\vspace{-.4mm}

\paragraph{Timing:}
Our model runs on average 21 ms per object instance. This is 14x faster than Polygon-RNN++ \cite{polygon-rnn++} and 1.4x faster than Curve GCN \cite{ling2019fast} which are the state of the arts.

\vspace{-2mm}
\section{Conclusion}
In this paper, we present PolyTransform, a novel deep architecture that combines the strengths of both prevailing segmentation approaches and modern polygon-based methods. We first exploit a segmentation network to generate a mask for each individual object. The instance mask is then converted into a set of polygons and serve as our initialization. Finally, a deforming network is applied to warp the polygons to better fit the object boundaries. We evaluate the effectiveness of our model on the Cityscapes dataset as well as a novel dataset that we collected.
Experiments show that our approach is able to produce precise, geometry-preserving instance segmentation that significantly outperforms the backbone model.
Comparing to the instance segmentation initialization, we increase the validation AP and boundary metric by up to 3.0 and 10.3 points, allowing us to achieve 1st place on the Cityscapes leaderboard. We also show that our model speeds up annotation by 35\%. Comparing to previous work on annotation-in-the-loop \cite{polygon-rnn++}, we outperform the boundary metric by 2.0\%. Importantly, our PolyTransform generalizes across various instance segmentation network.

{\small
\bibliographystyle{ieee_fullname}
\bibliography{egbib}
}
\end{document}